\RequirePackage{fix-cm}
\documentclass[twocolumn]{svjour3}          
\smartqed  
\usepackage{graphicx}
\usepackage[pagebackref=true,breaklinks=true,colorlinks,bookmarks=false,citecolor=blue,linkcolor=blue,urlcolor=blue]{hyperref}
\usepackage{amsmath,amssymb}
\usepackage{url}
\usepackage{paralist}
\usepackage{subfig}
\usepackage[countmax]{subfloat}
\usepackage{booktabs}
\usepackage{makecell}
\usepackage{textcomp}
\usepackage{gensymb}
\usepackage{pifont}
\newcommand{\cmark}{\ding{51}}%
\usepackage{float}
\usepackage{caption}
\usepackage[10pt]{moresize}
\usepackage{float}
\usepackage{xspace}
\usepackage{color}
\usepackage{multirow}
\usepackage{ifthen}

\long\def\ignorethis#1{}

\definecolor{gray}{rgb}{0.35,0.35,0.35}
\definecolor{MyBlue}{rgb}{0,0.2,0.8}
\definecolor{MyRed}{rgb}{0.8,0.2,0}
\definecolor{MyGreen}{rgb}{0.0,0.5,0.1}
\definecolor{MyGray}{rgb}{0.4,0.4,0.4}

\newlength\paramargin
\newlength\figmargin
\newlength\subfigmargin
\newlength\secmargin
\newlength\subsecmargin
\newlength\tabmargin
\newlength\eqmargin

\usepackage{array}
\newcolumntype{L}[1]{>{\raggedright\let\newline\\\arraybackslash\hspace{0pt}}m{#1}}
\newcolumntype{C}[1]{>{\centering\let\newline\\\arraybackslash\hspace{0pt}}m{#1}}
\newcolumntype{R}[1]{>{\raggedleft\let\newline\\\arraybackslash\hspace{0pt}}m{#1}}

\newcommand{\mpage}[2]
{
\begin{minipage}[t]{#1\linewidth}\centering
#2
\end{minipage}
}

\newcommand{\Paragraph}[1]
{\vspace{2mm} \noindent \textbf{#1}}

\def\ie{i.e.,~}
\def\eg{e.g.,~}

\def\etal{et~al.\xspace}

\newcommand{\figref}[1]{Fig.~\ref{fig:#1}}
\newcommand{\tabref}[1]{Table~\ref{tab:#1}}

\makeatletter 
  \newcommand\figcaption{\def\@captype{figure}\caption} 
  \newcommand\tabcaption{\def\@captype{table}\caption} 
\makeatother
\begin{document}

\title{Continuous and Diverse Image-to-Image Translation via 
Signed Attribute Vectors}

\author{Qi Mao \and
        Hung-Yu Tseng \and 
        Hsin-Ying Lee\and
        Jia-Bin Huang \and 
        Siwei Ma \and 
        Ming-Hsuan Yang
}


\institute{
 Qi Mao and Siwei Ma \at
             Institute of Digital Media, School of Electronics Engineering and Computer Science, Peking University, Haidian District, Beijing 100871, China. \\
              \email{\{qimao, swma\}@pku.edu.cn}
              \and
               Hung-Yu Tseng and Ming-Hsuan Yang \at
               Electrical Engineering and Computer Science, University of California at Merced, Merced, CA 95343 \\
              \email{\{htseng6, mhyang\}@ucmerced.edu}   
              \and
          Hsin-Ying Lee  \at
          Snap Research\\
          \email{hlee5@snap.com}
           \and
           Jia-Bin Huang \at
             Electrical and Computer Engineering, Virginia Tech, Blacksburg, VA 24060\\
             \email{jbhuang@vt.edu}
}
\date{Received: date / Accepted: date}

\maketitle

\begin{figure*}[ht]
\centering 
\includegraphics[width=\linewidth]{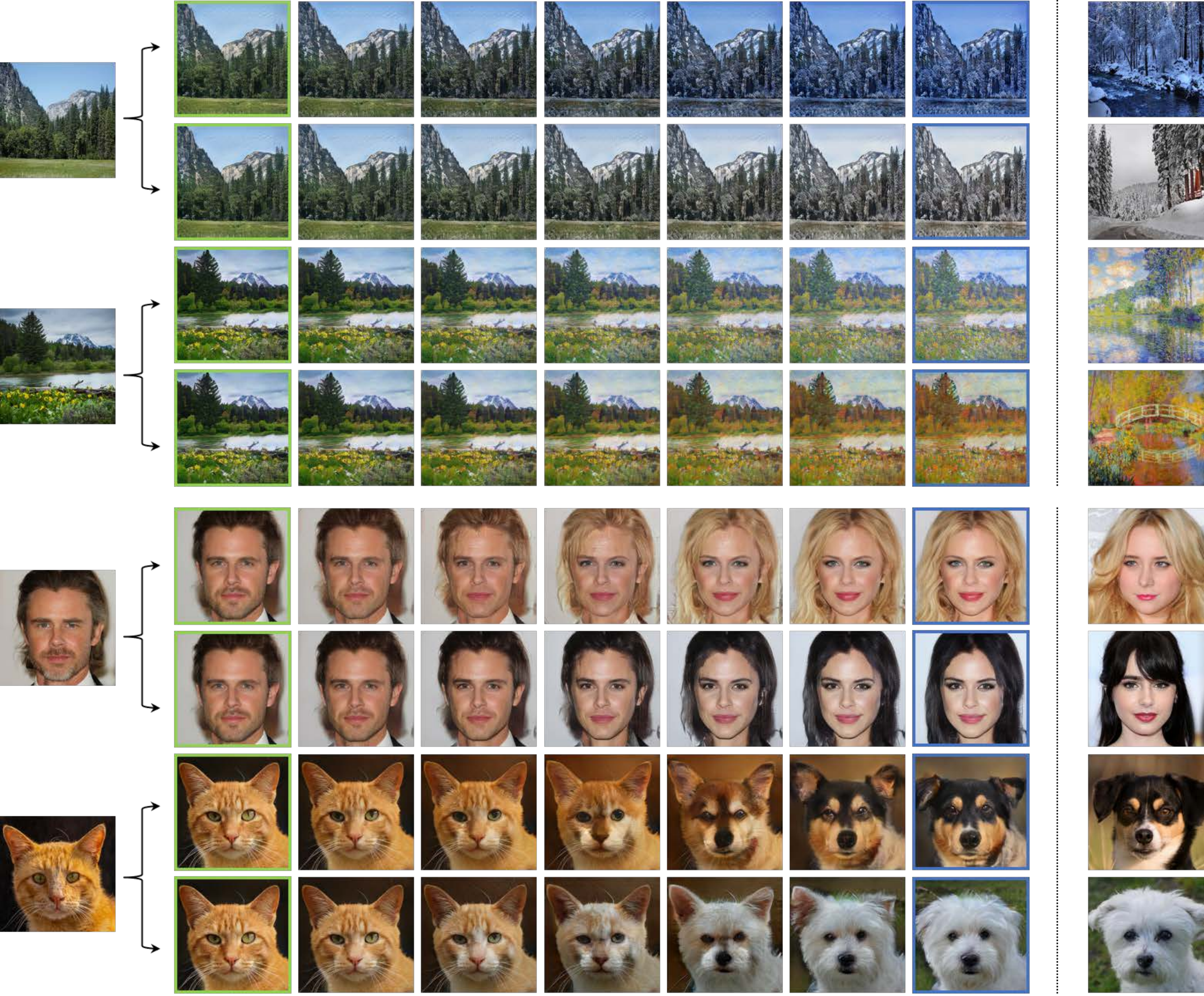}
     \mpage{0.1}{\textbf{Source}}\hfill\mpage{0.77}{\qquad Continuous interpolation results (Ours)
}\hfill\mpage{0.1}{\quad \textbf{Target}}
     
    \caption{\textbf{Continuous and diverse translation across domains.} 
    Our approach performs continuous and diverse I2I translation results. 
   On each row, we show continuous interpolation from the source image to the style of the target image, including 
    summer$\to$winter translation,
    photo$\to$Monet translation,
    male$\to$female translation, and cat$\to$dog translation.
    Green and blue bounding boxes denote generated images of the source and target domains, respectively.
    }

    \label{fig:teaser}
\end{figure*}
\begin{table*}[t]
    \caption{\textbf{Comparisons of I2I translation networks.} Our model renders continuous translation on diverse mapping paths across multiple domains.}
    \label{tab:related_work}
    \centering
    \renewcommand\tabcolsep{2.5pt} 
    \scriptsize
    \begin{tabular}{l cccccccc} 
        \toprule
        Methods & MUNIT \cite{huang2018multimodal} & DRIT \cite{lee2018diverse} & DMIT \cite{yu2019multi}&DRIT++ \cite{lee2019drit++} & StarGAN-v2 \cite{choi2020stargan} & DLOW \cite{gong2019dlow} & FUNIT \cite{liu2019few} & Ours \\
        \midrule
        Multi-modal & \cmark & \cmark & \cmark &\cmark &\cmark & - & \cmark &\cmark\\
        Multi-domain & - &  - &\cmark &\cmark & \cmark & - & \cmark &\cmark\\
        Continuous translation & - & -& - & - &- & \cmark & \cmark &\cmark\\
        \bottomrule
    \end{tabular}
\end{table*}

\begin{abstract}
Recent image-to-image (I2I) translation algorithms focus on learning the mapping from a source to a target domain.
However, the continuous translation problem that synthesizes intermediate results between two domains has not been well-studied in the literature.
Generating a smooth sequence of intermediate results bridges the gap of two different domains, facilitating the morphing effect across domains.
Existing I2I approaches are limited to either intra-domain or deterministic inter-domain continuous translation.
In this work, we present an effectively signed attribute vector, which enables \emph{continuous} translation on \emph{diverse} mapping paths \emph{across} various domains.
In particular, we introduce a unified attribute space shared by all domains that utilize the sign operation to encode the domain information, thereby allowing the interpolation on attribute vectors of different domains.
%
To enhance the visual quality of continuous translation results, we generate a trajectory between two sign-symmetrical attribute vectors and leverage the domain information of the interpolated results along the trajectory for adversarial training. 
We evaluate the proposed method on a wide range of I2I translation tasks.
Both qualitative and quantitative results demonstrate that the proposed framework generates more high-quality continuous translation results against the state-of-the-art methods.
\end{abstract}

\section{Introduction}
\label{sec:introduction}
Image-to-Image (I2I) translation \cite{isola2017image} aims to learn the mapping function between different visual domains.
It can be applied to a wide range of tasks such as semantic image synthesis \cite{wang2018high,park2019SPADE}, photo enhancement \cite{zhu2017unpaired}, style transfer \cite{lee2019drit++,gong2019dlow}, season transfer \cite{huang2018multimodal,lee2018diverse,yu2019multi}, domain adaptation~\cite{hoffman2018cycada,chen2019crdoco}, and object transfiguration \cite{wu2019transgaga,choi2018stargan,liu2019few,choi2020stargan}.
Given an image of the source domain, we can render it into an image with the target domain's style.
%
However, it remains challenging to generate smooth and continuous translated images for existing I2I methods. 
As shown in \figref{teaser}, continuous translation enables applications such as image morphing. 
Furthermore, modeling the intermediate results facilitates a better understanding of the translation process between the two domains.

%
Existing I2I translation approaches mainly address multi-modal \cite{zhu2017toward,huang2018multimodal,lee2018diverse}, multi-domain \cite{choi2018stargan,anoosheh2018combogan,liu2019few} translation, or both \cite{yu2019multi,lee2019drit++,choi2020stargan}.
However, these methods are not effective in rendering continuous translated images across domains. 
While recent translation frameworks~\cite{zhu2017toward,huang2018multimodal,lee2018diverse,lee2019drit++,choi2020stargan} based on domain-specific attribute (style) space can generate continuous translation by performing linear interpolation between different attribute vectors, such schemes are limited to \emph{intra}-domain due to separate attribute spaces of different domains.
To handle continuous I2I translation, the DLOW model~\cite{gong2019dlow} generates intermediate results by taking an additional interpolated domain label as input.
However, this method can only produce a deterministic translation path given an image and an interpolated domain label.
%

In this paper, we focus on \emph{continuous} image-to-image translation and generate \emph{diverse} translation paths (multi-modal) across various visual domains (multi-domain) in a single model. 
There exist two challenges. 
%
%
First, previous approaches such as MUNIT \cite{huang2018multimodal} and DRIT \cite{lee2019drit++} adopt separate attribute spaces for different domains, thus cannot model the continuous variation across domains.
Second, there are no ground-truth intermediate samples for learning continuous translation results in-between two domains.


%
To address the issues mentioned above, we propose a novel I2I translation framework based on signed attribute vectors (SAVs).
We disentangle images into the content and the attribute representations extracted by a content encoder and an attribute encoder.
To enable inter-domain continuous translation, we introduce a \emph{unified} attribute space containing domain-specific attributes of all domains.
We consider each attribute dimension from the prior Gaussian distribution as independent and identically distributed random variables and draw samples.  
Then, the \emph{sign} operation is applied to make the values of the attributes in a particular domain positive and those in other domains negative.
The proposed SAVs and the content representations are fed into the generator to synthesize corresponding domain images.
Furthermore, we adopt the maximum mean discrepancy (MMD) constraint to align the distribution of SAVs with the attribute encoder embedding distribution.
Thus, the attribute representation can either be sampled from the signed attribute space or extracted from the given reference image. 

There are two advantages of the proposed method.
First, it facilitates continuous translation across various domains within a unified attribute space.
Second, owing to the sign information, we propose a translation trajectory between the SAV of the source domain and its \emph{sign-symmetrical} attribute vector of the target domain.    
We leverage the domain information on interpolated results along the trajectory during the training and apply the adversarial loss to enhance the quality.
\figref{teaser} shows the effectiveness of the proposed method for generating continuous and diverse translation results across domains.

The main contributions of this work are summarized as follows:

\begin{compactitem}
\item We propose a simple yet effective SAV-based I2I framework to construct a unified attribute space for all domains, enabling continuous and diverse translation paths across various visual domains.
\item 
We design a sign-symmetrical operation to create a translation trajectory between two domains during training.
By leveraging the domain information of intermediate results along the trajectory, we apply the adversarial loss to enhance the quality. 
\item
Extensive experiments validate that the proposed method can synthesize continuous and diverse translation results on a wide range of I2I translation tasks.
Qualitative and quantitative results demonstrate that the proposed framework generates more high-quality continuous translation results against the state-of-the-art approaches.
\end{compactitem}


\section{Related Work}
\label{sec:related}
\vspace{\paramargin}
\subsection{Image-to-image Translation.}
Image-to-Image (I2I) translation \cite{isola2017image} aims to learn the mapping of images between different domains.
%
%
Isola \etal \cite{isola2017image} propose Pix2Pix to address the problem with paired data. 
Numerous I2I methods \cite{zhu2017unpaired,liu2017unsupervised,kim2017learning} exploit the idea of cycle-consistency to train the model with unpaired data.
However, these approaches can only perform one-to-one (\ie uni-modal) mapping translation.
%

Recent methods achieve one-to-many mapping from different perspectives: multi-modal \cite{zhu2017toward,huang2018multimodal,lee2018diverse}, multi-domain \cite{choi2018stargan,anoosheh2018combogan,liu2019few} translation, or both \cite{yu2019multi,lee2019drit++,choi2020stargan}.
Nevertheless, these schemes focus on generating images of the target distribution, ignoring the continuous translation process that produces intermediate results across domains.

Several recent approaches aim to model the intermediate state between two domains for continuous translation.
The DLOW method \cite{gong2019dlow} introduces an additional interpolated domain variable, which facilitates mapping the source image to the intermediate domain.
Lira \etal~\cite{lira2020ganhopper} design multi-hops in one generator to gradually transform the source image to the target domain.
However, both approaches are built upon the uni-modal translation model Cycle-GAN \cite{zhu2017unpaired}.
The generator only synthesizes one deterministic continuous translation path to the target domain for a source image.

To achieve multi-modality, MUNIT \cite{huang2018multimodal} and DRIT \cite{lee2018diverse} disentangle images into domain-invariant content representations and domain-specific attribute representations.
However, due to the separate construction of domain-specific attribute space of different domains, they can only perform continuous interpolation within the intra-domain.  
Several schemes, such as DRIT++ \cite{lee2019drit++}, DMIT \cite{yu2019multi}, and StarGAN-v2 \cite{choi2020stargan}, integrate the attribute (style) representations and domain labels in a single framework for multi-modal and multi-domain translations.
Nevertheless, they still do not perform well in continuous translation across multiple domains.
The attribute encoder and the generator of DRIT++ \cite{lee2019drit++} require an additional domain label as input, which prevents interpolation on attribute vectors from different domains. 
DMIT \cite{yu2019multi} disentangles images into the domain-invariant content space, the domain-shared style space, and the domain-specific attribute space represented by domain labels.
Directly interpolating vectors of domain-shared style space does not lead to continuous translations from the source domain to the target one.
Despite that StarGAN-v2 \cite{choi2020stargan} introduces the unified style encoder to encode style information of different domains, the design of multiple embedding branches still separates attribute space of different domains. 
Applying interpolation straightforwardly to style vectors of different domains does not generate smooth translated images, as shown in our experimental results.  
In contrast, we present a unified attribute representation that contains domain-specific attributes from all domains.
The SAVs are then proposed utilizing the sign operation to embed the domain information, facilitating \emph{continuous} translation on \emph{diverse} mapping paths \emph{across} various domains.

Numerous few-shot I2I translation methods~\cite{liu2019few,saito2020coco}, \eg FUNIT~\cite{liu2019few}, construct a unified class encoder to learn a class-specific style code for different classes.
These approaches can achieve continuous translation by interpolating two style codes of different classes.
However, without explicit constraint on the latent space, we note that the interpolated results depend on the number of translation classes and the batch size during training, as shown in our experiments.
In contrast, owing to the SAVs, we create a translation trajectory between two domains using two sign-symmetrical attribute vectors.
By leveraging the domain information of intermediate results along the trajectory, the adversarial loss can further enhance interpolated images.
Table \ref{tab:related_work} summarizes the differences between recent unsupervised I2I translation frameworks.

\subsection{Image Synthesis by Varying Attributes}
Numerous image synthesis methods have been developed by varying the underlying attributes.
For style transfer, Kotovenko \etal~\cite{kotovenko2019content} vary the art style of input photo from one artist to another, \eg from Cezanne to Van Gogh, for continuous image synthesis.
For head reenactment, Burkov~\etal~\cite{burkov2020neural} learn the latent pose representation and synthesize images with varying views by interpolating two pose vectors.

Several facial attributes continuous manipulation approaches can continuously manipulate one specific attribute, such as the smile attribute on face images, by varying the value of the annotated attribute label.
The Fader Networks \cite{lample2017fader} method uses binary attribute labels for training.
The model can then be generalized to use continuous float values to manipulate the specific facial attribute strength during inference.
The RelGAN \cite{wu2019relgan} scheme proposes a relative-attribute-based approach for facial attribute editing.
Guided by an interpolation discriminator, the framework learns to interpolate a specific attribute during training, leading to a smoother continuous facial attribute manipulation.
Instead of linearly interpolating the annotated attribute labels, the recent approach HomoInterpGAN \cite{chen2019homomorphic} embeds images into latent representations by an encoder and learns an interpolator network to interpolate the latent features. 
The interpolator network is then trained with a homomorphic loss to manipulate latent representations consistent with annotated attributes.
However, since these approaches require fine-grained binary annotations during the training stage, they are not directly applicable to the continuous I2I task, which only provides few domain labels.
Recently, several approaches explore interpretable attributes in the latent space of pre-trained generative adversarial network (GAN) models~\cite{shen2020interpreting,voynov2020unsupervised} (\ie GAN inversion) for image synthesis.

In contrast to existing methods, we aim to continuously manipulate the domain-specific attributes from the source to the target domains. 
The proposed framework learns the domain-specific attributes directly from the data for I2I translation.

\subsection{Image Morphing.}
Image morphing \cite{wolberg1998image} aims to change from an image to another through a seamless transition.
Existing approaches \cite{wolberg1998image} usually accomplish the task via multiple steps: determining the corresponding mapping between the images on specific feature space \cite{liao2014automating}, applying a 2D geometric transformation to warp the image for retaining geometric alignment on the feature space, interpolating the color space to blend the texture.

With the advances of GANs, numerous methods \cite{abdal2019image2stylegan,abdal2019image2stylegan++} manipulate the latent space of state-of-the-arts GAN models such as StyleGAN \cite{abdal2019image2stylegan}, StyleGAN-v2 \cite{abdal2019image2stylegan++}, and BigGAN \cite{brock2018large}, which have been demonstrated to be effective for image morphing. 
In particular, they project images into these pre-trained GAN models' latent space and then perform the morphing effect by interpolating latent vectors of different images.

Unlike interpolating latent features embedded by pre-trained GANs, we learn to disentangle images into the content and attribute representations.
By interpolating attribute vectors of different domains, we generate morphing effects to the target domain while preserving the source image's content.


\begin{figure*}[!t]
\begin{center}
\includegraphics[width=.95\linewidth]{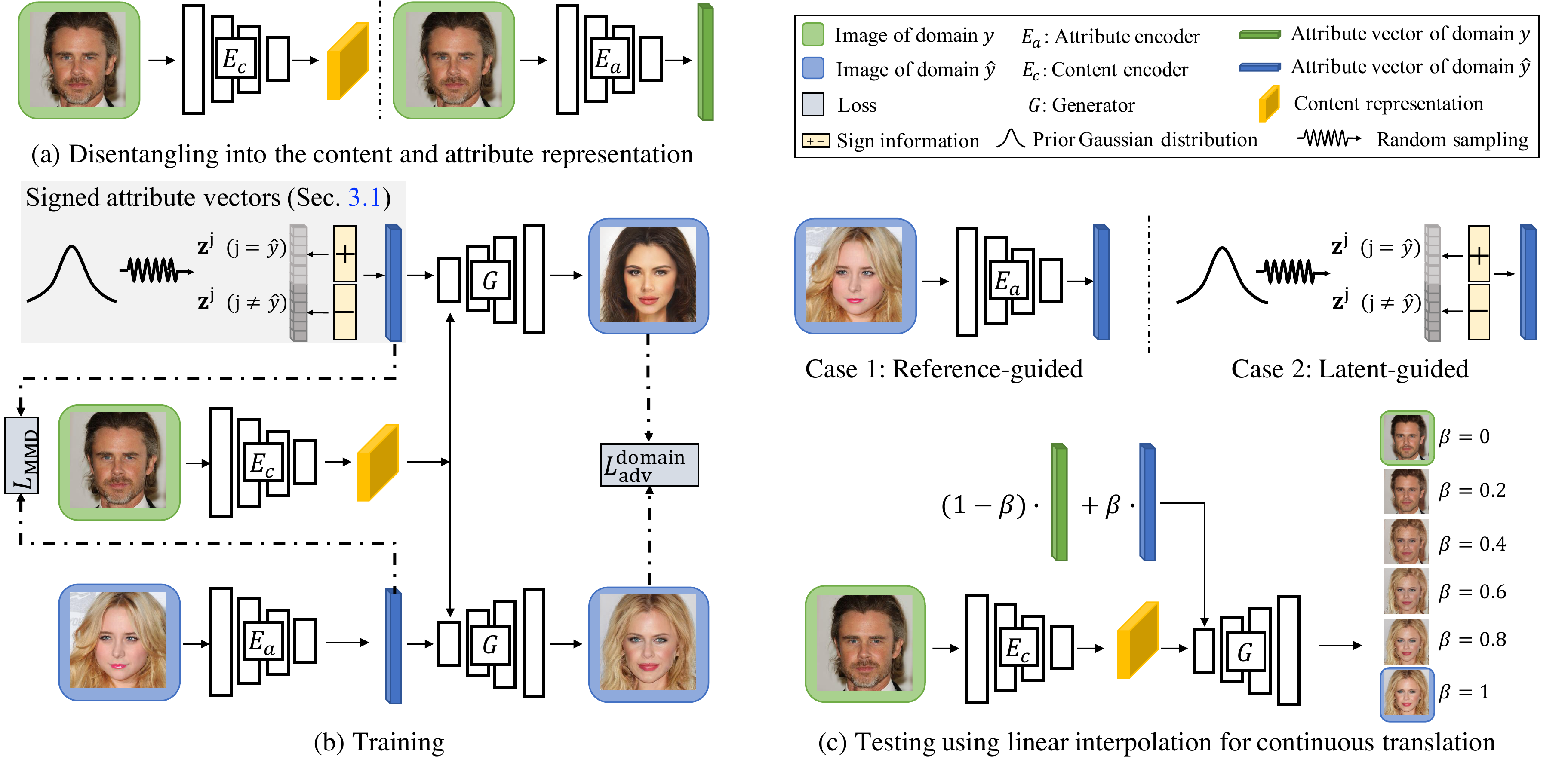}
\end{center}
\caption{\textbf{Overview.} 
(a) An image is disentangled into the content and attribute representations by a unified content encoder $E_c$ and a unified attribute encoder $E_a$.
(b) We present the male$\to$female translation and design a unified attribute space containing attributes of all domains. 
The sign information is incorporated into the attribute vector to describe domain membership.
Given a target domain~$\hat{y}$, we enforce the domain-specific attributes $\mathbf{z}^{j}(j=\hat{y})$ positive and others $\mathbf{z}^{j}(j\neq\hat{y})$ negative, which forms the SAVs (Sec. \ref{sec:3-1}).
On the other hand, we extract the attribute vector from the reference image of the domain~$\hat{y}$, whose distribution is aligned with the SAVs of the domain~$\hat{y}$ using the MMD constraint.
Finally, $G$ synthesizes the translated images using the content representation extracted from the source image and the target attribute vector.
(c) In the testing phase, we extract the target attribute vector of a reference image (case 1: reference-guided) or randomly generate an SAV of the target domain (case 2: latent-guided).
The continuous translation is realized by interpolating the extracted attribute vector of the source image and the target attribute vector.
}

\label{fig:framework}
\end{figure*}

\section{Proposed Method}
\label{sec:proposed}
%
Our goal is to learn continuous and diverse I2I translation across visual domains while preserving the domain-invariant content.
As illustrated in Fig. \ref{fig:framework}(a), given an image $\mathbf{x}$, we use a content encoder $E_{c}$ to obtain the domain-invariant content representation $\mathbf{c}$, and a unified attribute encoder $E_{a}$ to extract the attribute vector $\mathbf{z}$.
We can then generate continuous translation by interpolating two different attribute vectors.
However, existing approaches \cite{huang2018multimodal,lee2018diverse,lee2019drit++,choi2020stargan} are more effective for intra-domain interpolation due to separate attribute spaces for different domains.
To enable \emph{inter-}domain continuous translation, we propose a unified attribute space shared by all domains, denoted as
\begin{equation}
\small
\begin{gathered}
\hspace{4mm}
  \mathbf{z} = [z^1_1,z^1_2,\cdots,z^1_d,z^2_1,z^2_2,\cdots,z^2_d,\cdots,z^N_1,z^N_2,\cdots,z^N_d], \\
  \mathbf{z}\in \mathbb{R}^{d\cdot N},
\end{gathered}
\end{equation}
where $d$ is the attribute vector's dimension for each domain, and $ N $ represents the number of visual domains.
In the following, we present a sign operation to encode the domain information into the unified attribute vector and the framework's training strategies.

\subsection{Signed Attribute Vectors}
\label{sec:3-1}
In this work, we encode the domain information into the unified attribute space.
The prior of the translation between two domains lies in one domain that has some more \emph{prominent} attributes than the other domain. 
Given a source image $\mathbf{x}$ of the domain $y$, we assume that the attribute values corresponding to the domain $y$ should be large, while those from other domains should be relatively small.
For instance, the beard is usually longer on male faces compared to female faces.
Therefore, we propose to use the sign operation to formulate this assumption.

\begin{figure}[!t]
\centering
	\begin{minipage}{0.235\textwidth}
    \subfloat[\footnotesize{Sign-symmetrical points}]{
    \includegraphics[width=\linewidth]{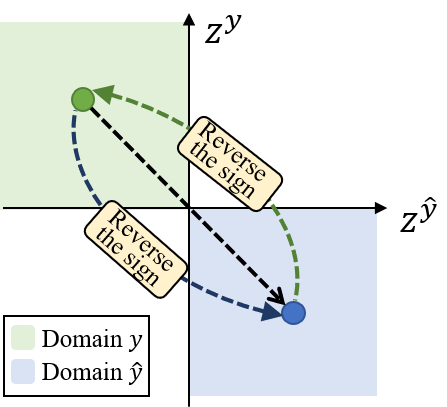}
    }
    \end{minipage}
    \begin{minipage}{0.235\textwidth}
	\subfloat[\footnotesize{$\beta<0.5$: Source domain}]{
    \includegraphics[width=\linewidth]{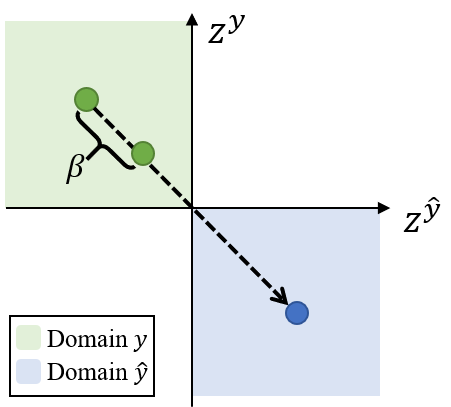}
    }
    \end{minipage}

	\begin{minipage}{0.235\textwidth}
    \subfloat[\footnotesize{$\beta=0.5$: Intermediate state}]{
    \includegraphics[width=\linewidth]{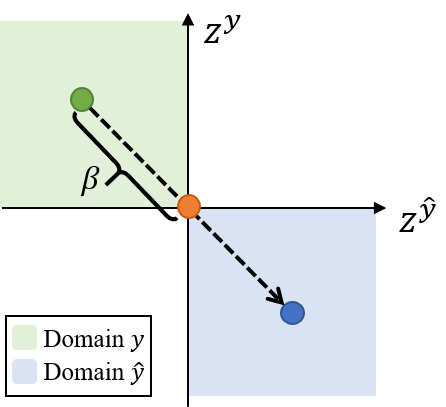}
    }
    \end{minipage}
    \begin{minipage}{0.235\textwidth}
    \subfloat[\footnotesize{$\beta>0.5$: Target domain}]{
    \includegraphics[width=\linewidth]{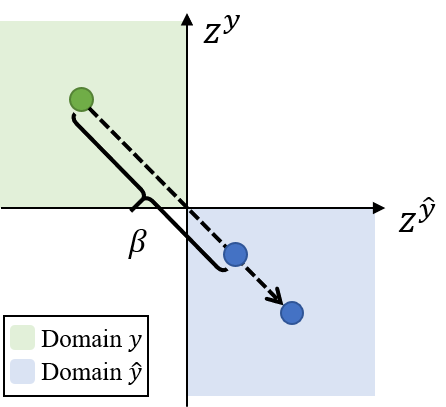}
    }
    \end{minipage}
\caption{\textbf{Illustration of interpolation between two sign-symmetrical points.}
Using a 2D-plane, we show that the signed attribute vector contains attributes of two domains, and each domain has one attribute dimension. 
(a) The sign-symmetrical attribute vector is obtained by reversing the sign of the attributes in the source and target domains. 
(b) When $\beta < 0.5$, the interpolated point lies in the source domain.
(c) When $\beta = 0.5$, the interpolated point is an intermediate state.
(d) When $\beta > 0.5$, the interpolated point belongs to the target domain.
}
\label{fig:framework2}
\end{figure}
First, we sample a vector $\small{\mathbf{z}^p\in\mathbb{R}^{d\cdot N}}$ from a prior distribution.
Each attribute dimension of the vector is \textit{i.i.d.} sampled from the prior Gaussian distribution $\small{\mathcal{N}(0,1)}$.  
For a domain label $y$, we use the sign operation to compute the SAVs:
\begin{equation}
    \small
\hspace{8mm}
    \mathbf{z}^s=\mathcal{O}_y(\mathbf{z}^p)\hspace{5mm}\mathbf{z}^p\backsim \mathcal{N}(\mathbf{0},\mathbf{I}),  y\in\{1...N\}.
\end{equation}
Specifically, the sign operation $\mathcal{O}_y$ makes the attribute values $\small{\{z^{j=y}_i\}_{i=1}^d}$ of domain $y$ positive, and those for other domains $\small{\{z^{j\neq y}_i\}_{i=1}^d}$ negative by,
\begin{equation}
\begin{gathered}
    \small
\hspace{12mm}
    \mathcal{O}_y(\mathbf{z}^p)= [
-|z^1_1|,-|z^1_2|,\cdots,
-|z^1_d|, \cdots,  \\
+|z_1^{y}|, +|z_2^{y}|, \cdots, +|z_d^{y}|, \cdots, \\ -|z^N_1|,-|z^N_2|,\cdots,-|z^N_d|].
\end{gathered}
\end{equation}

We show an example of the proposed sign operation on an attribute vector of target domain $\hat{y}$ in the gray block of \figref{framework}(b).
Then, the SAV $\mathbf{z}^s$ can be applied to align the distribution of the unified attribute vector $\mathbf{z}$ extracted by the attribute encoder $E_a$ using the MMD \cite{zhao2017infovae} constraint:
\begin{equation}
\small
\begin{gathered}
\hspace{3mm}
   \mathcal{L}_{\mathrm{MMD}} = \mathbb{E}_{p(\mathbf{z}^s),p(\mathbf{z})}[k(\mathbf{z}^s,\mathbf{z})]+\mathbb{E}_{q(\mathbf{z}^s),q(\mathbf{z})}[k(\mathbf{z}^s,\mathbf{z})]\\
   -2\mathbb{E}_{p(\mathbf{z}^s),q(\mathbf{z})}[k(\mathbf{z}^s,\mathbf{z})],
  \end{gathered}
\end{equation}
where $k$ is the Gaussian kernel $\small{k(\mathbf{z}^s,\mathbf{z})=e^{-\frac{\|\mathbf{z}^s-\mathbf{z}\|^2}{2\sigma^2}}}$.

As presented in Fig. \ref{fig:framework}(b), the content representation extracted from the source image along with the attribute representation, which is either sampled by the signed attribute space or extracted from the reference target image, are fed into the generator to synthesize the translated images of target domain $\hat{y}$. 
We use a multi-task discriminator $D$ \cite{mescheder2018training,liu2019few,choi2020stargan} with multi-branch outputs for all domains during the training stage to ensure that the translated images belong to the corresponding domain.
For target domain $\hat{y}$, the corresponding branch $D_{\hat{y}}$ is learned by the domain adversarial loss,
\begin{equation}
     \small
    \hspace{4mm}
    \label{eq:gan}
\begin{aligned}
    \mathcal{L}_{\mathrm{adv}}^{\mathrm{domain}}&=\mathbb{E}_{\mathbf{x},\mathbf{x}_{\hat{y}}}[\log D_{\hat{y}}(\mathbf{x}_{\hat{y}})+\log (1-D_{\hat{y}}(G(\mathbf{c},\mathbf{z}))], \\
       \mathrm{or} &=\mathbb{E}_{\mathbf{x},\mathbf{x}_{\hat{y}},\mathbf{z}^s}[\log D_{\hat{y}}(\mathbf{x}_{\hat{y}})+\log (1-D_{\hat{y}}(G(\mathbf{c},\mathbf{z}^s))], 
\end{aligned}
\end{equation}
where $\mathbf{x}_{\hat{y}}$ is the real image of target domain $\hat{y}$, $\small\mathbf{c} = E_c(\mathbf{x})$ is the content representation of the source image, and the attribute vector can be derived from $\mathbf{z}=E_a(\mathbf{x}_{\hat{y}})$ or $ \mathbf{z}^s=\small\mathcal{O}_{\hat{y}}(\mathbf{z}^{p})$.

We embed only the sign information in the attribute vectors sampled from a prior distribution in this work.
For the unified attribute vectors extracted by the attribute encoder $E_a$, the sign information is learned via the domain adversarial loss $\mathcal{L}_{\mathrm{adv}}^{\mathrm{domain}}$ and the MMD constraint $\mathcal{L}_{\mathrm{MMD}}$.
Furthermore, we adopt the \textbf{style reconstruction loss} \cite{Johnson2016Perceptual} to ensure the style consistency of translated images with reference images:
\begin{equation}
\begin{aligned}
\hspace{5mm}
   \mathcal{L}_{\mathrm{style}}&=\mathbb{E}_{\mathbf{x},\mathbf{x}_{\hat{y}}}[|\mathrm{Gram}(\phi(G(E_c(\mathbf{x}), E_a(\mathbf{x}_{\hat{y}}))) \\
   - &\mathrm{Gram}(\phi(\mathbf{x}_{\hat{y}}))\|_1],
  \end{aligned}
\end{equation}
where $\mathrm{Gram}$ is gram matrix, and $\phi$ is embedding feature space. 
In particular, we use ReLU3$\_$1 VGG features.

With such a simple sign operation on SAVs, we construct a unified attribute space incorporating the domain information, thus allowing continuous translation across domains.
Images are translated by interpolating the extracted attribute vector of a source image and a target attribute vector of the target domain.
As illustrated in \figref{framework}(c), the target attribute vector can be obtained from a reference image (case 1: reference-guided) or a randomly sampled SAV of the target domain (case 2: latent-guided).

\subsection{Improving Quality by Sign-Symmetrical Attribute Vectors}
\label{sec:3-2}
Thanks to the unified attribute space, we can interpolate the attribute vectors of two different domains.
However, improving the quality of intermediate results remains a challenge.
Due to the lack of real interpolation samples, we cannot directly apply the domain adversarial loss in Eq. \ref{eq:gan} to ensure the interpolated images' realism.
Therefore, we propose two sign-symmetrical attribute vectors to create a continuous translation trajectory for conducting the domain adversarial training.

Given an SAV $\mathbf{z}^{s}$ from source domain $y$, we reverse the attribute sign of source domain $y$ and that of target domain $\hat{y}$.
As an example shown in \figref{framework2}(a), it produces a \emph{sign-symmetrical} attribute vector $\mathbf{z}_{\mathrm{sym}}$ for the target domain.
We formulate the process as
\begin{equation}
\small
\begin{gathered}
\mathbf{z}_{\mathrm{sym}}=\mathcal{O}^{r}(\mathbf{z}^{s})=[
-|z^1_1|,-|z^1_2|,\cdots, -|z^1_d|,\cdots, \\
+|z_1^{\hat{y}}|, +|z_2^{\hat{y}}|, 
\cdots, +|z_d^{\hat{y}}|, \cdots, 
-|z_1^{y}|, -|z_2^{y}|, ..., -|z_d^{y}|, \cdots, \\
-|z^N_1|,-|z^N_2|,\cdots,-|z^N_d|].
\end{gathered}
\end{equation}

For the translated image $\hat{\mathbf{x}}$ using the sign-symmetrical attribute vector $\mathbf{z}_{\mathrm{sym}}$, we first apply the domain adversarial loss to assure its domain membership belongs to target domain $\hat{y}$ and name it as the \textbf{reverse sign domain adversarial loss} $\mathcal{L}_{\mathrm{adv}}^{\mathrm{rvs}}$:
\begin{equation}
    \small
    \label{eq:gan_rvs}
    \hspace{10mm}
  \mathcal{L}_{\mathrm{adv}}^{\mathrm{rvs}}=\mathbb{E}_{\mathbf{x},\mathbf{x}_{\hat{y}},\mathbf{z}^{s}}[\log D_{\hat{y}}(\mathbf{x}_{\hat{y}})+\log (1-D_{\hat{y}}(\hat{\mathbf{x}}))],
\end{equation}
where $\small{\hat{\mathbf{x}}=G(\mathbf{c}, \mathbf{z}_{\mathrm{sym}})}$, and $\mathbf{c} = E_c(\mathbf{x})$.

Since traversing these two vectors $\mathbf{z}^{s}$ and $\mathbf{z}_\mathrm{sym}$ forms a continuous translation trajectory across domains, we leverage this path during the training stage.
Specifically, we sample the interpolation coefficient from a uniform distribution, \ie $\beta \in (0,1)$, and conduct linear interpolation on the two sign-symmetrical vectors $\mathbf{z}^s$ and $\mathbf{z}_\mathrm{sym}$ in domain $y$ and $\hat{y}$, respectively.
The interpolated attribute vector can be formulated as $\mathbf{z}^\mathrm{i} = (1-\beta) \cdot \mathbf{z}^{s} +  \beta \cdot \mathbf{z}_{\mathrm{sym}}$.
Combining a content representation $\mathbf{c}$, we generate the interpolated translated result $\mathbf{x}^\mathrm{i} = G(\mathbf{c},\mathbf{z}^\mathrm{i})$.
Based on the sign information of the attribute vector, there exist three cases of $\mathbf{x}^\mathrm{i}$:
\begin{compactitem}
\item
1) When $\beta < 0.5$, the interpolated attribute vector still locates in domain $y$, as illustrated in \figref{framework2}(b).
Therefore, the generated interpolated image belongs to domain $y$.
\item
2) When $\beta > 0.5$, the interpolate attribute vector lies in target domain $\hat{y}$, as demonstrated in \figref{framework2}(d).
At this time, the interpolated image belongs to domain $\hat{y}$.
\item
3) When $\beta = 0.5$, this is the intermediate state, as shown in \figref{framework2}(c).
In this case, we regard the interpolated result indistinguishable from neither domain $y$ nor domain $\hat{y}$.
Compared to $\beta \in (0,0.5)$ and $\beta \in (0.5,1)$, this point is much less sampled. 
\end{compactitem}
As a result, we can apply the domain adversarial loss to images generated along this trajectory to ensure the quality of continuous translation results, which is defined as the \textbf{interpolated domain adversarial loss} $\mathcal{L}_{\mathrm{adv}}^{\mathrm{interp}}$ and summarized as follows
\begin{equation}
\scriptsize
\mathcal{L}_{\mathrm{adv}}^{\mathrm{interp}}=\left\{
\begin{array}{lr}
\mathbb{E}_{\mathbf{x},\mathbf{z}^s,\beta}[\log D_{y}(\mathbf{x})+\log (1-D_{y}(\mathbf{x}^\mathrm{i}))] & {\beta < 0.5,}\\
\mathbb{E}_{\mathbf{x},\mathbf{x}_{\hat{y}},\mathbf{z}^s,\beta}[\frac{1}{2} \log D_{y}(\mathbf{x})+\frac{1}{2} \log (1-D_{y}(\mathbf{x}^\mathrm{i})) +\\
\frac{1}{2}\log D_{\hat{y}}(\mathbf{x}_{\hat{y}})+\frac{1}{2}\log (1-D_{\hat{y}}(\mathbf{x}^\mathrm{i}))]&{\beta = 0.5,} \\
\mathbb{E}_{\mathbf{x},\mathbf{x}_{\hat{y}},\mathbf{z}^s,\beta}[\log D_{\hat{y}}(\mathbf{x}_{\hat{y}})+\log (1-D_{\hat{y}}(\mathbf{x}^\mathrm{i}))] &{\beta > 0.5.}\\
\end{array} \right.
\end{equation}

Although we only exploit this specific interpolation path during training,
the model is generalized to ensure interpolation results between any two attribute vectors at the inference stage. 


\begin{figure*}[htbp]
\centering 
\subfloat[Interpolation on the Photo2Artwork dataset (Photo$\to$Monet, Photo$\to$Van Gogh, and Photo$\to$Ukiyo-e)]{
\mpage{1}{
\includegraphics[width=1\linewidth]{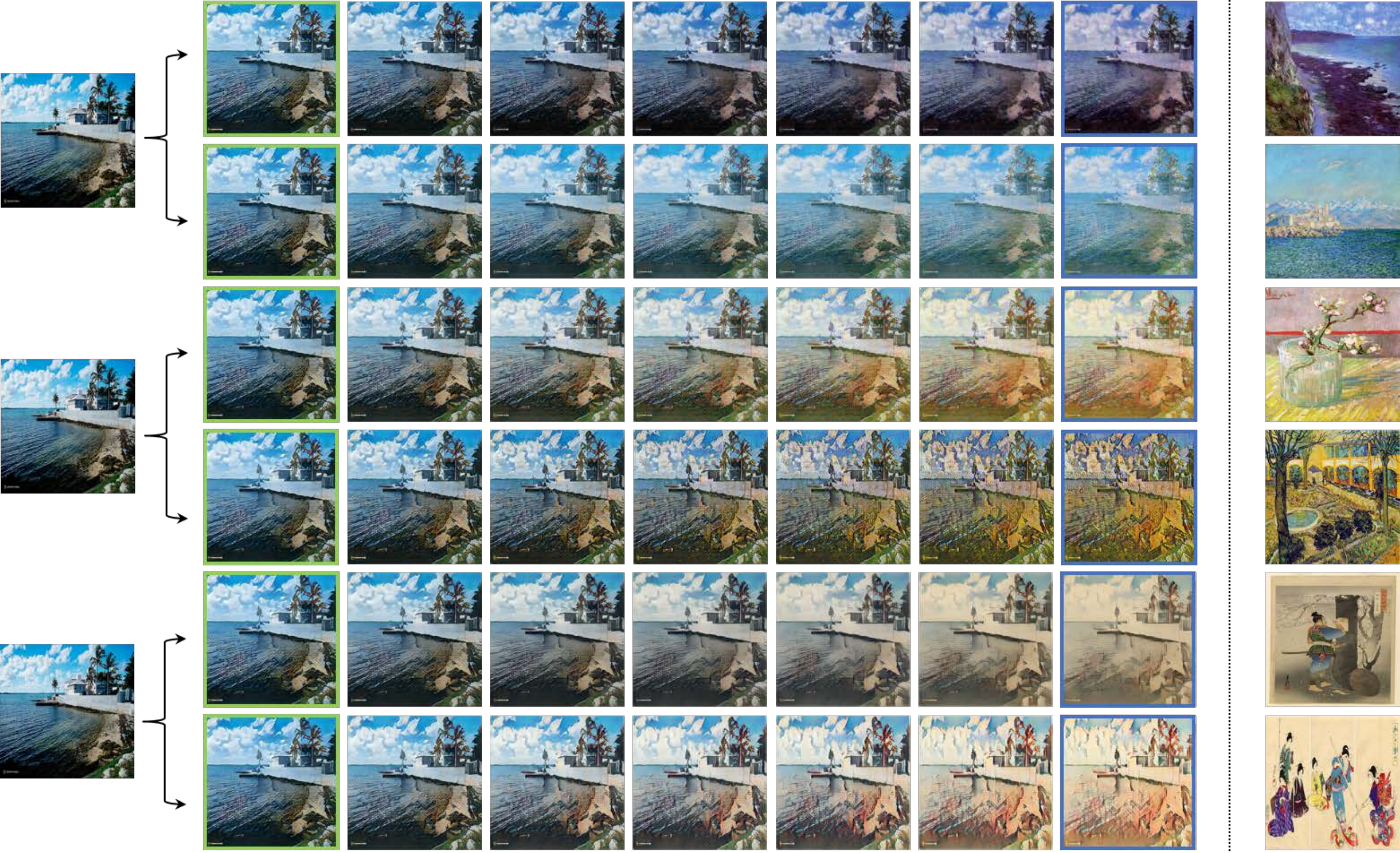}
 \mpage{0.1}{\textbf{Source}}\hfill\mpage{0.77}{\qquad Continuous translation results
}\hfill\mpage{0.1}{\quad \textbf{Target}}}}

\subfloat[Interpolation on the AFHQ dataset (Cat$\to$Dog and Cat$\to$Wildlife)]{
\mpage{1}{
\includegraphics[width=1\linewidth]{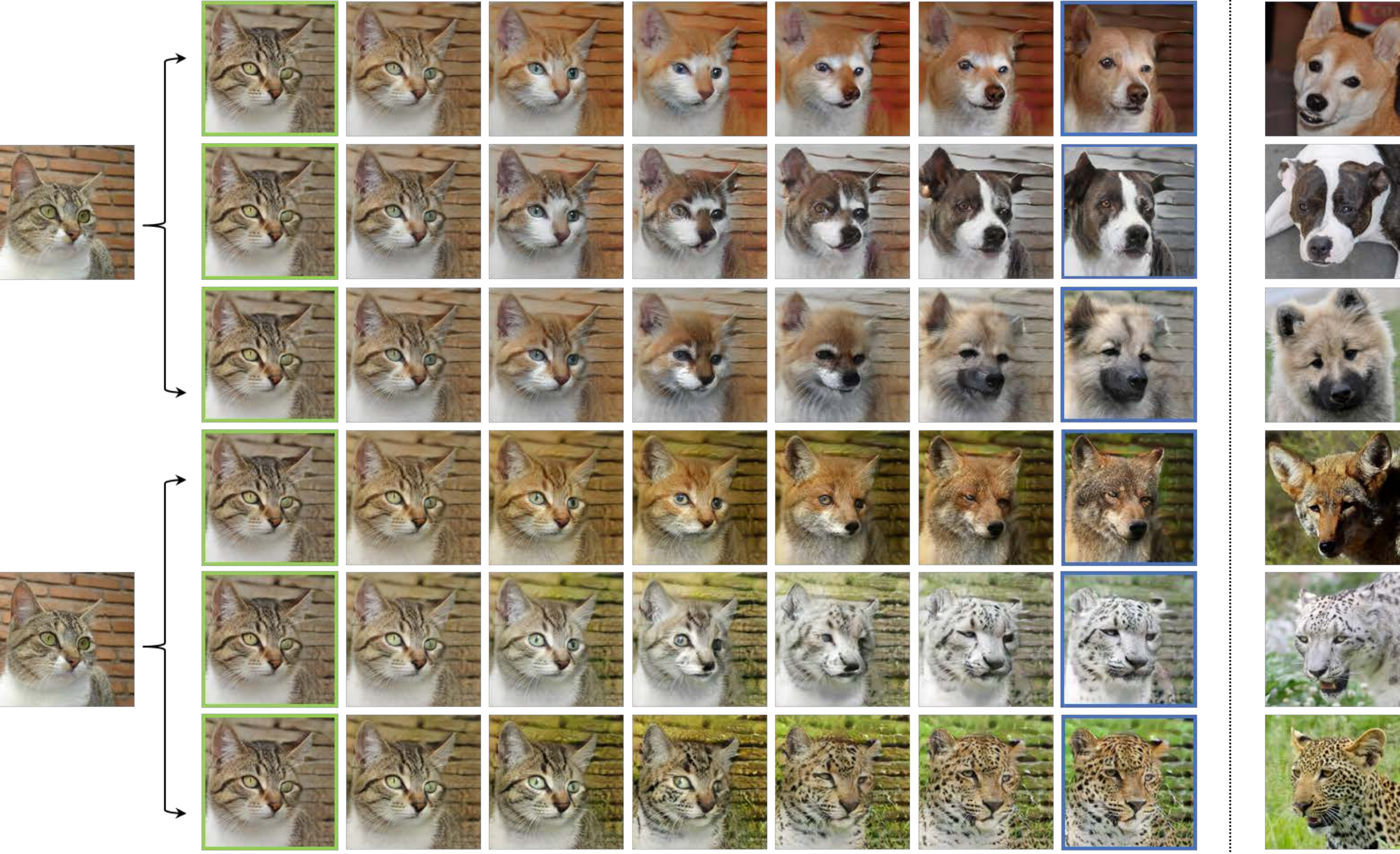}
 \mpage{0.1}{\textbf{Source}}\hfill\mpage{0.77}{\qquad Continuous translation results
}\hfill\mpage{0.1}{\quad \textbf{Target}}}
     } 
\caption{\textbf{Reference-guided continuous and diverse translation results.} 
(a) Translation results on natural scenes where there are no dominant objects in the scenes.
(b) Translations results on animals where one dominant object is in the scene. 
On each row, we show continuous translation results from the source to the target domains using the attribute vector extracted from a reference image.
Green and blue bounding boxes denote generated images of the source and target domains, respectively.
 %
}
\label{fig:result1}
\end{figure*}

\subsection{Other Loss Objectives}
\label{sec:3-3}
\vspace{\paramargin}
In addition to the above loss functions, we also apply several loss objectives commonly used in I2I translation approaches to train the proposed model.

\Paragraph{Content adversarial loss.}
To further disentangle the content and attribute representations, we adopt the content discriminator $D_{c}$ \cite{lee2018diverse,lee2019drit++,yu2019multi} to distinguish the content representations belong to different domains.
On the other hand, the content encoder $E_{c}$ aims to generate the content representations fool the content discriminator $D_{c}$.
Then, the content adversarial loss is defined by
\begin{equation}
\small
\begin{aligned}
\mathbb{L}_{\mathrm{adv}}^{\mathrm{content}} &=\mathbb{E}_{\mathbf{x},\mathbf{x}_{\hat{y}}}[\frac{1}{2}\log D_{c}(E_{c}(\mathbf{x})+\frac{1}{2}\log (1-D_{c}(E_{c}(\mathbf{x}))\\
+ & \frac{1}{2}\log D_{c}(E_{c}(\mathbf{x}_{\hat{y}})+\frac{1}{2}\log (1-D_{c}(E_{c}(\mathbf{x}_{\hat{y}}))].
\end{aligned}
\end{equation}

\Paragraph{Cycle-consistency loss.} To preserve the consistency of domain-invariant characteristics of generated images, we impose the cycle-consistency loss \cite{choi2020stargan,zhu2017unpaired}, 
\begin{equation}
 \small
 \begin{aligned}
 \hspace{15mm}
\mathcal{L}_{\mathrm{1}}^{\mathrm{cc}}&=\mathbb{E}_{\mathbf{x},\mathbf{x}_{\hat{y}}}[\|\mathbf{x} - G(E_c(\hat{\mathbf{x}}), E_a(\mathbf{x}))]\|_1],\\
\mathrm{or} & = \mathbb{E}_{\mathbf{x},\mathbf{z}_{s}}[\|\mathbf{x} - G(E_c(\hat{\mathbf{x}}), E_a(\mathbf{x}))]\|_1],
\end{aligned}
\end{equation}
where $\hat{\mathbf{x}}= G(E_c(\mathbf{x}), E_a(\mathbf{x}_{\hat{y}}))$ or $\hat{\mathbf{x}}= G(E_c(\mathbf{x}),\mathcal{O}_{\hat{y}}(\mathbf{z}^{p}))$.

\Paragraph{Self-reconstruction loss.} We reconstruct the original image \cite{lee2018diverse,lee2019drit++} using the encoded content representation and attribute representation as
\begin{equation}
\small
\hspace{15mm}
\mathcal{L}_{\mathrm{1}}^{\mathrm{recon}}=\mathbb{E}_{\mathbf{x}}[\|G(E_c(\mathbf{x}), E_a(\mathbf{x})) - \mathbf{x} \|_1].
\end{equation}

\vspace{\paramargin}
\Paragraph{Latent regression loss} \cite{zhu2017toward,lee2018diverse,lee2019drit++,yu2019multi,choi2020stargan} is adopted to further encourage the invertible mapping between generated images and the signed attribute space.
We reconstruct the signed attribute vector $\mathbf{z}^s$ as
\begin{equation}
\hspace{12mm}
    \mathcal{L}_{1}^{\mathrm{latent}}= \mathbb{E}_{\mathbf{x},\mathbf{z}^{s}}[| E_{a}(G(E_c(\mathbf{x}), \mathbf{z}^s)) -\mathbf{z}^s\|_1].
\end{equation}

\vspace{\paramargin}
\Paragraph{Mode seeking loss.} To alleviate the mode collapse problem and improve the diversity of generated images.
We introduce another SAV $\mathbf{z}_2^s$ to calculate the mode seeking loss \cite{mao2019mode} as
\begin{equation}
\hspace{2mm}
    \max \ \mathcal{L}_{\mathrm{ms}}= \mathbb{E}_{\mathbf{x},\mathbf{z}_1^{s},\mathbf{z}_2^{s}}[\frac{\|G(E_c(\mathbf{x}), \mathbf{z}_1^{s}) -G(E_c(\mathbf{x}), \mathbf{z}_2^{s})\|_1}{\|\mathbf{z}_1^{s} -\mathbf{z}_2^{s}\|_1}].
\end{equation}

The objective function of our framework is
\begin{equation}
\small
\begin{split}
 \min\limits_{G,E_c,E_a}  \max\limits_{D,D_c} &\lambda_\mathrm{adv}^{\mathrm{content}}\mathcal{L}_{\mathrm{adv}}^\mathrm{content} + \lambda_\mathrm{adv}^{\mathrm{domain}}\mathcal{L}_{\mathrm{adv}}^{\mathrm{domain}}\\
&+\lambda_\mathrm{adv}^{\mathrm{rvs}}\mathcal{L}_{\mathrm{adv}}^{\mathrm{rvs}}+ \lambda_\mathrm{adv}^{\mathrm{interp}}\mathcal{L}_{\mathrm{adv}}^{\mathrm{interp}},\\
 \min\limits_{G,E_c,E_a}  &\lambda_{\mathrm{MMD}}\mathcal{L}_{\mathrm{MMD}}+\lambda_{\mathrm{style}}\mathcal{L}_{\mathrm{style}}+\lambda_1^{\mathrm{cc}}\mathcal{L}^{\mathrm{cc}}_{1}\\ &+\lambda_1^{\mathrm{recon}}\mathcal{L}^{\mathrm{recon}}_{1} +\lambda_1^{\mathrm{latent}}\mathcal{L}^{\mathrm{latent}}_{1}+\lambda_{\mathrm{ms}}\frac{1}{\mathcal{L}_{\mathrm{ms}}},
\end{split}
\end{equation}
where the term $\lambda_{\ast}$ controls the importance of each loss function.
%

\section{Implementation Details}
\label{sec:implementation}
The proposed model is implemented in Pytorch \cite{paszke2017automatic}, and the source code and pre-trained models are available at \url{https://github.com/HelenMao/SAVI2I}. 
%

\Paragraph{Datasets}
We evaluate the proposed method on four representative datasets, including the style translation and shape-variation translation tasks.
For style translation, 
the Yosemite \cite{zhu2017unpaired} dataset includes the summer and the winter two domains.
The Photo2Artwork dataset \cite{zhu2017unpaired} contains the photo, Monet, Van Gogh, and Ukiyo-e domains. 
For shape-variation translation, 
the CelebA-HQ \cite{karras2017progressive} dataset in which
we split the male and female domains for translation.
The AFHQ \cite{choi2020stargan} dataset consists of animal faces with the cat, dog, and wildlife domains.

\Paragraph{Network Architecture.}
The proposed model consists of a content encoder $E_c$, an attribute encoder $E_a$, a generator $G$, a discriminator $D$, and a content discriminator $D_c$. 
We set the size of an attribute vector to $\mathbf{z}\in \mathbb{R}^{8\cdot N} $, where $N$ represents the number of visual domains in the dataset.
To better fuse the attributes, we feed the attribute vectors into a fusing network $F$ with three-layer MLP before feeding them into the generator.
Since the style translation tasks require more content preservation than shape-variation translation tasks, we adopt different network architecture choices for these two tasks. 
More details on the network architectures can be found in the supplementary material.

\Paragraph{Training Process.}
The resolution of 
each image is $256\times256$ pixels for all the experiments.
We adopt the following hyper-parameters for the training in all the experiments:
$\lambda_\mathrm{adv}^{\mathrm{content}}=1$, $\lambda_\mathrm{adv}^{\mathrm{domain}}=1$, $\lambda_\mathrm{adv}^{\mathrm{rvs}}=1$, $\lambda_\mathrm{adv}^{\mathrm{interp}}=1$, 
$\lambda_{\mathrm{style}}=1$,
$\lambda_1^{\mathrm{cc}}=10$,
$\lambda_1^{\mathrm{recon}}=10$,
$\lambda_1^{\mathrm{latent}}=10$, 
$\lambda_{\mathrm{ms}}=1$.
For AHFQ, we employ $\lambda_{\mathrm{MMD}}=10$, and $\lambda_{\mathrm{MMD}}=1$ for other datasets.
We use the batch size of $1$ as well as the Adam \cite{kingma2014adam} optimizer with a learning rate of $10^{-4}$ and exponential decay rates $(\beta_1, \beta_2)=(0, 0.99)$. 
We adopt the non-saturating adversarial loss \cite{goodfellow2014generative} with $R_1$ regularization \cite{mescheder2018training} using $\gamma=10$ for style translation tasks, and $\gamma=1$ for shape-variation translation tasks.
All of the models are trained on two NVIDIA Tesla-P$100$ GPUs with 16GB memory.

\section{Experiments}
\label{sec:experiments}
\subsection{Continuous and Diverse Image-to-Image Translation}
We present diverse continuous translation paths from the source domain to the target one.
An input image $I_s$ in the source domain can be continuously translated to various images $I_{t_1}, I_{t_2}, \dots, I_{t_N}$ in the target domain.
As illustrated in Fig. \ref{fig:framework}(c), we first embed an image of the source domain to obtain the content representation and the source attribute vector.
Then, we compute a target attribute vector by either extracting a reference image sampled from the target domain (\emph{reference-guided}) or randomly generate an SAV of the target domain (\emph{latent-guided}).
We apply linear interpolation of the source and target attribute vectors to generate continuous translation results.
\figref{result1} shows examples of reference-guided continuous and diverse translation results for two different scenarios: one with holistic views and the other one with dominant objects in the scenes. 
For the first scenario, the translation task is similar to style transfer, where the holistic views are considered (i.e., look and feel).
For the second scenario, the translation task focuses on varying the dominant object's shape and texture in each scene. 
Our model learns to continuously vary the source's attributes to the target domains in both translation tasks. 
%
More results can be found in the supplementary material.

\subsection{Comparisons with the State-of-the-arts}
We present qualitative and quantitative evaluations with the state-of-the-art approaches on the CelebA-HQ and AFHQ datasets.

\begin{figure*}[!thbp]
\centering
   \begin{minipage}{1\textwidth}
    \subfloat[Qualitative comparisons ]{
\includegraphics[width=1\linewidth]{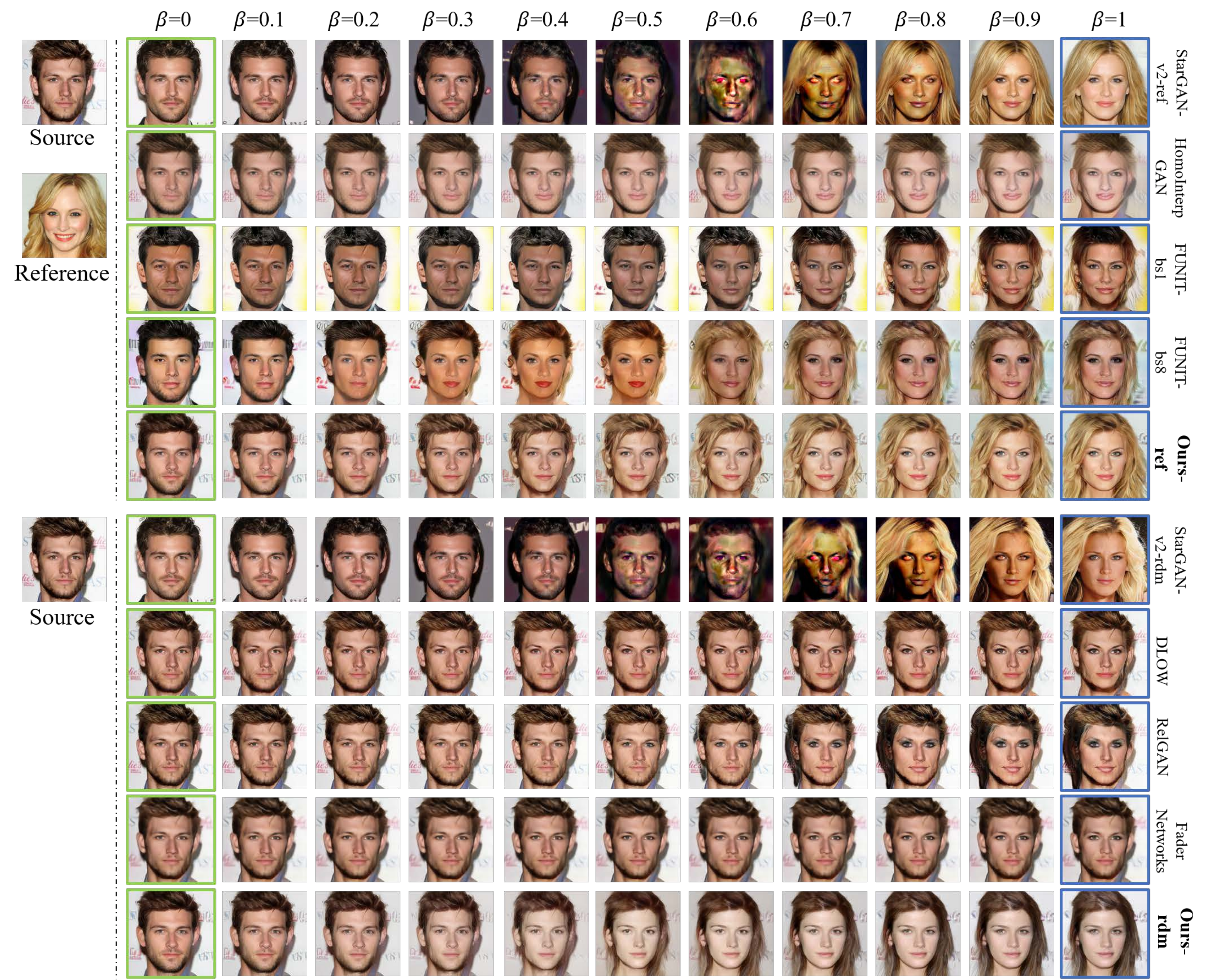}}
\end{minipage}

   \begin{minipage}{1\textwidth}
    \subfloat[Quantitative comparisons]{
    \includegraphics[width=\linewidth]{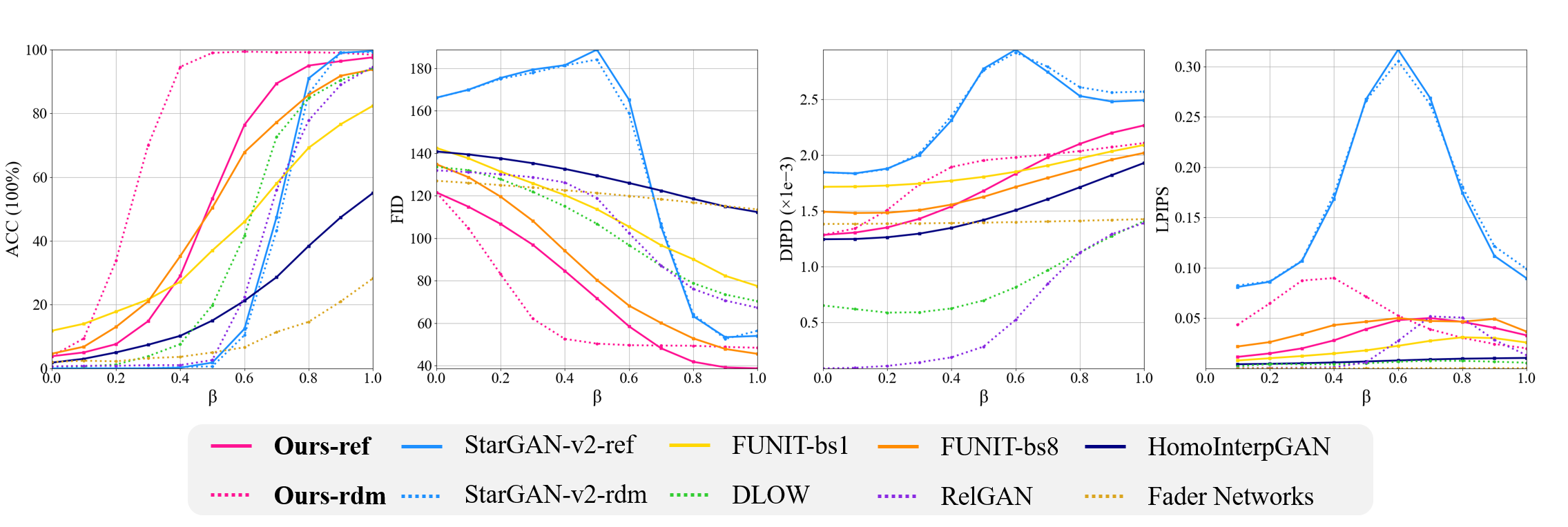}
    }
    \end{minipage}
    \caption{\textbf{Translation from male $\to$ female images. 
     }
  (a) Translation based on reference images (top);
    Translation based on interpolated domain labels or randomly sampled latent vectors (bottom).
   Green and blue bounding boxes denote generated images of the source and target domain, respectively.
(b) From left to right, the y-axis of each sub-figure are target domain translation ACC (the larger, the better), target domain FID (the smaller, the better), DIPD (the smaller, the better), and LPIPS between two adjacent interpolated images (the smaller, the better).
    Each curve is plotted under different $\beta$ values.
    Solid lines indicate methods using reference images.
    Dash lines denote approaches using interpolated domain labels or randomly sampled latent vectors of the target domain.}
    %
    
    \label{fig:compare_result1}
\end{figure*}
\subsubsection{State-of-the-art Methods.}
We evaluate the proposed method against the state-of-the-art models, including:
\begin{itemize}
    \item I2I translation approaches: StarGAN-v2 \cite{choi2020stargan}, FUNIT \cite{liu2019few}, and DLOW \cite{gong2019dlow}.
    \item Attribute variation schemes: HomoInterpGAN \cite{chen2019homomorphic}, RelGAN \cite{wu2019relgan}, and Fader Networks \cite{lample2017fader}.
\end{itemize}
For the I2I translation approaches, we adopt the pre-trained StarGAN-v2 model provided by the authors.
Although the FUNIT model is originally designed to address the few-shot setting of I2I translation, we use the same training protocols to train the FUNIT model and adopt two settings for performance evaluation: training the model with the batch size of 1 (the same as ours) and the model with the batch size of 8.
%
For multi-domain translation, the DLOW method mixes styles of multiple target domains and does not perform well. 
As such, we train three models on the AFHQ dataset for translation between any two domains.
Instead of using fine-grained attribute annotations,
we only use domain labels to train I2I models. 
For fair companions, we use the models and codes provided by their authors.

\subsubsection{Evaluation Metrics}
For quantitative evaluation, we use four widely-used metrics to assess interpolated images under different $\beta$ values:
\begin{itemize}
    \item Target domain translation accuracy (ACC)~\cite{chen2019homomorphic,liu2019few}. We measure the percentage that interpolated images belong to the target domain, \ie
\begin{equation}
    ACC = \frac{\sum_{i=1}^N \delta [\mathcal{C}(\mathbf{x}_{i})==y^{target}]}{N},
\end{equation}
where $N$ is the total number of interpolated images, $\mathbf{x}_{i}$ is the $i$-th example of interpolated images, $y^{target}$ is the target domain label, $ \mathcal{C}(\cdot)$ is a domain classifier that predicts domain label of $\mathbf{x}_{i}$, and $\delta(\cdot)$ is a function that outputs 1 if predicted label equals to the target domain label, and 0 otherwise.
We use the ResNet-50 binary classifier in our experiments.

    \item Target domain Fr\'{e}chet inception distance (FID)~\cite{heusel2017gans}.
    We compute the FID score between interpolated images and real images of the target domain.
    
    \item Domain-invariant perceptual distance (DIPD)~\cite{liu2019few}.%
    We calculate the $L_2$ distance between two normalized VGG Conv5~\cite{simonyan2015very} features extracted from the interpolated image and the source image.
    
    \item Learned perceptual image patch similarity (LPIPS) score~\cite{zhang2018unreasonable} between two adjacent interpolated images.
\end{itemize}

The target domain translation ACC and FID scores measure whether translated results successfully change into the target domain and the degree of image realism against the target domain.
The DIPD score evaluates whether interpolated images preserve the domain-invariant features of source images.
The variation of the LPIPS score between two adjacent interpolated images can be considered an indicator of translation smoothness.

\begin{figure*}[!thbp]
\centering
   \begin{minipage}{1\textwidth}
    \subfloat[Qualitative comparisons]{
\includegraphics[width=1\linewidth]{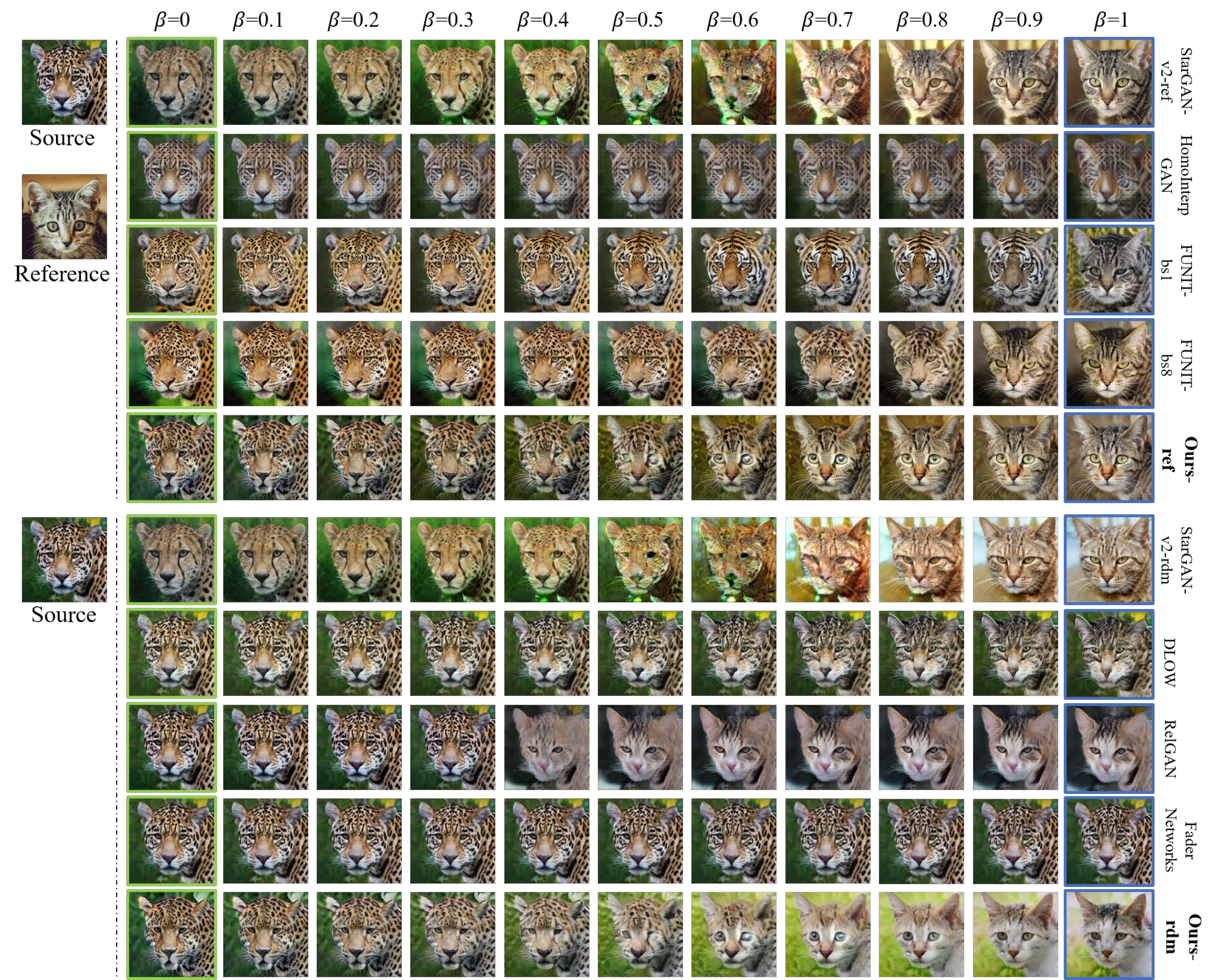}}
\end{minipage}

   \begin{minipage}{1\textwidth}
    \subfloat[Quantitative comparisons]{
    \includegraphics[width=\linewidth]{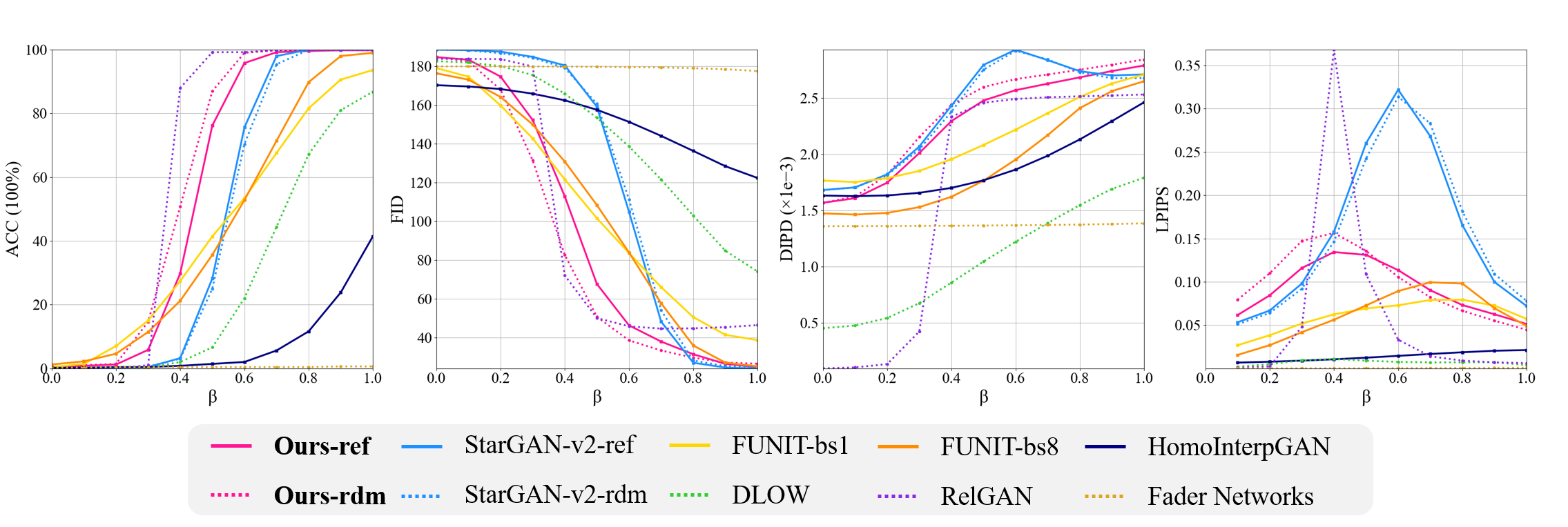}
    }
    \end{minipage}

    \caption{\textbf{Translation from wildlife $\to$ cat images. 
     }
(a) Translation based on reference images (top);
    Translation based on interpolated domain labels or randomly sampled latent vectors (bottom).
   Green and blue bounding boxes denote generated images of the source and target domain, respectively.
(b) From left to right, the y-axis of each sub-figure are target domain translation ACC (the larger, the better), target domain FID (the smaller, the better), DIPD (the smaller, the better), and LPIPS between two adjacent interpolated images (the smaller, the better).
    Each curve is plotted under different $\beta$ values.
    Solid lines indicate methods using reference images.
    Dash lines denote approaches using interpolated domain labels or randomly sampled latent vectors of the target domain.}      
    \label{fig:compare_result2}
\end{figure*}

\subsubsection{Evaluation Protocol}
We randomly select $500$ examples from the test set of each domain for evaluation.
The interpolated coefficient $\beta$ continuously takes the value at an interval of $0.1$ from $0$ to $1$.
When generating continuous results, the FUNIT~\cite{liu2019few} and HomoInterpGAN~\cite{chen2019homomorphic} methods use reference images in the target domain as guidance. 
The DLOW~\cite{gong2019dlow}, RelGAN~\cite{wu2019relgan}, and Fader Networks~\cite{lample2017fader} interpolate discrete domain labels to generate intermediate results. 
The StarGAN-v2 and proposed schemes use target attribute vectors either extracted from reference images or randomly sampled from the target domain's latent space for interpolation.

We feed the source and target image into the class encoder of the FUNIT method \cite{liu2019few} and interpolate the class codes according to the interpolated coefficients to generate interpolated results.
The HomoInterpGAN scheme \cite{chen2019homomorphic} embeds the source and target image into the unified latent feature space by the encoder.
We adjust the value of the control vector $\textbf{v} \in [0,1]^{c\times1}$ of the domain branch to control the interpolation results.
For the StarGAN-v2 model, we feed the source and target image into the style encoder~\cite{choi2020stargan} and obtain style vectors from the source and target domain branches.
Then, we apply linear interpolation on these two style vectors.
In addition, we can randomly sample a latent vector from the Gaussian distribution and feed it into the mapping network to acquire the target domain's style vector from the corresponding domain branch~\cite{choi2020stargan}.
We then apply interpolation between source and target style vectors.
The DLOW method~\cite{gong2019dlow} uses interpolated domain labels as additional inputs.
For the RelGAN approach~\cite{wu2019relgan}, we construct relative-attribute-vectors regarding domain labels and multiply it with interpolated coefficients.
The Fader Networks scheme generates attributes based on domain labels~\cite{lample2017fader} and applies linear interpolation between the source and target domain attributes.

\begin{table*}[!t]
	\centering
	\caption{\textbf{User preference scores.} The numbers denote the percentage of users who prefer the proposed method over the comparative approach.}
	\vspace{-1mm}
	\begin{tabular}{@{}cccc ccc} 
	    \toprule
		Setting & \multicolumn{3}{c}{CelebA-HQ}&\multicolumn{3}{c}{AFHQ}
		\\  \cmidrule(lr){2-4} \cmidrule(lr){5-7}
	 & Realism & Smoothness & Preference & Realism & Smoothness & Preference  \\
Ours-ref vs. StarGAN-v2-ref	& 86.25 & 98.75 & 96.25 & 77.50& 82.50 & 83.30	 \\ 
Ours-ref vs. HomoInterpGAN & 96.25 & 87.50 & 96.25 & 96.67 & 89.17 & 96.67	 \\ 
Ours-ref vs. FUNIT-bs1&	 82.50 & 85.00 & 83.75 & 63.33 & 62.50 & 64.17 \\ 		
Ours-ref vs. FUNIT-bs8&	 83.75 & 80.00 & 85.00 & 70.83 & 75.83 & 72.50 \\
\midrule
Ours-rdm vs. StarGAN-v2-rdm & 88.75 & 97.50 & 95.00 & 67.50 & 72.50 & 72.50	 \\ 
Ours-rdm vs. DLOW &	 97.50 & 93.75 & 97.50 & 91.67 & 80.00 & 91.67 \\ 
Ours-rdm vs. RelGAN & 93.75& 93.75 & 95.00 & 94.17 & 95.00& 95.00	 \\ 
Ours-rdm vs. Fader Networks & 95.00 & 87.50 &96.25 & 98.33 & 87.50 & 98.33	 \\ 
		\bottomrule 
	\end{tabular}
	\label{tab:user_study}
\end{table*}

\subsubsection{Experimental Results Analysis}
\figref{compare_result1}(a) and \figref{compare_result2}(a) show  translation results by the evaluated methods.
In \figref{compare_result1}(b) and \figref{compare_result2}(b), we compute the target domain translation ACC and FID values using all interpolated images and calculate the average DIPD and LPIPS scores of all interpolated images at each $\beta$ value to plot the curves ``ACC vs. $\beta$'', ``FID vs. $\beta$'', ``DIPD vs. $\beta$'', and ``LPIPS vs. $\beta$''.
In these plots, ``Ours-ref'' and ``StarGAN-v2-ref'' indicate continuous translation using target attribute vectors extracted from reference exemplars; 
``Ours-rdm'' and ``StarGAN-v2-rdm'' represent continuous translation using latent vectors randomly sampled from the target domain;
``FUNIT-bs1'' as well as ``FUNIT-bs8'' denote the FUNIT model trained with the batch size of $1$ and $8$.

\Paragraph{I2I translation approaches}.
The StarGAN-v2 method does not generate intermediate results well due to the separate attribute spaces for different domains.
The curves of ``DIPD vs. $\beta$'' and ``LPIPS vs. $\beta$'' thus have an abrupt mutation and form a peak when $\beta$ is 0.6, as described in \figref{compare_result1}(b) and \figref{compare_result2}(b).
In contrast, our method can generate smooth intermediate results across domains with the proposed SAV.

For I2I translation of face images, the FUNIT model cannot capture target images' style well with the class code when trained with only two classes, as illustrated in \figref{compare_result1}(a).
Thus, \figref{compare_result1}(b) demonstrates that the target domain translation ACC and FID values do not perform as well as the proposed model.
The interpolated images of the FUNIT-bs8 method with a large batch size have higher realism scores. 
However, without any explicit constraint in the latent space, the FUNIT-bs8 model does not generate a smooth transition, as shown in \figref{compare_result1}(a). 
The FUNIT model achieves better continuous interpolation in animal faces translation when more training classes are used.
Nevertheless, \figref{compare_result2}(b) shows that the proposed model performs better against it in terms of ACC and FID scores.  
The DLOW method can translate images to the target domains (the ACC score is larger than $50\%$ when $\beta$ is larger than $0.5$). 
However, the translated results contain only local variations (\eg makeup) and do not exhibit the hairstyle in \figref{compare_result1}(a).
The ACC and FID scores achieved by the DLOW method are lower than those by the proposed model, as shown in \figref{compare_result1}(b) and \figref{compare_result2}(b).

\Paragraph{Attribute variation schemes}.
The synthesized images by the RelGAN approach do not undergo smooth transition. 
For example, the leopard ($\beta=0.3$) changes into a cat ($\beta =0.4$) in \figref{compare_result2}(a).
The spike ($\beta =0.4$) of the curve in \figref{compare_result2}(b) ``LPIPS vs. $\beta$''  also shows the RelGAN approach is not able to translate images smoothly.
On the other hand, although the HomoInterpGAN and Fader Networks nearly have no variation in the curve of ``LPIPS vs. $\beta$'', they cannot continuously translate the source image into the target domain (\ie the ACC score is less than $50\%$ when $\beta$ is larger than $0.5$), as shown in \figref{compare_result1}(b) and \figref{compare_result2}(b).

Overall, our approach synthesizes images with desired attributes such as hairstyle, makeup, and skin-color in the male$\to$female translation.
The proposed method achieves the best ACC and FID scores when $\beta$ is larger than $0.5$ among all the evaluated schemes, as shown in \figref{compare_result1}(b) and \figref{compare_result2}(b). 
In terms of smooth I2I translation, the LPIPS scores increase and decrease steadily. 
Unlike the StarGAN-v2 and RelGAN methods, the curves of LIPIS scores by the proposed method do not contain any spikes.
Furthermore, the trends of curves in ``ACC vs. $\beta$'', ``FID vs. $\beta$'', and ``DIPD vs. $\beta$'' are consistent without any abrupt changes when increasing the $\beta$ value.
These results demonstrate that the proposed model can accomplish both high-quality interpolation and smooth transition.
%

\begin{figure*}[!t]
    \centering
  
   \begin{minipage}{1\textwidth}
    \subfloat[Qualitative comparisons]{
\includegraphics[width=1\linewidth]{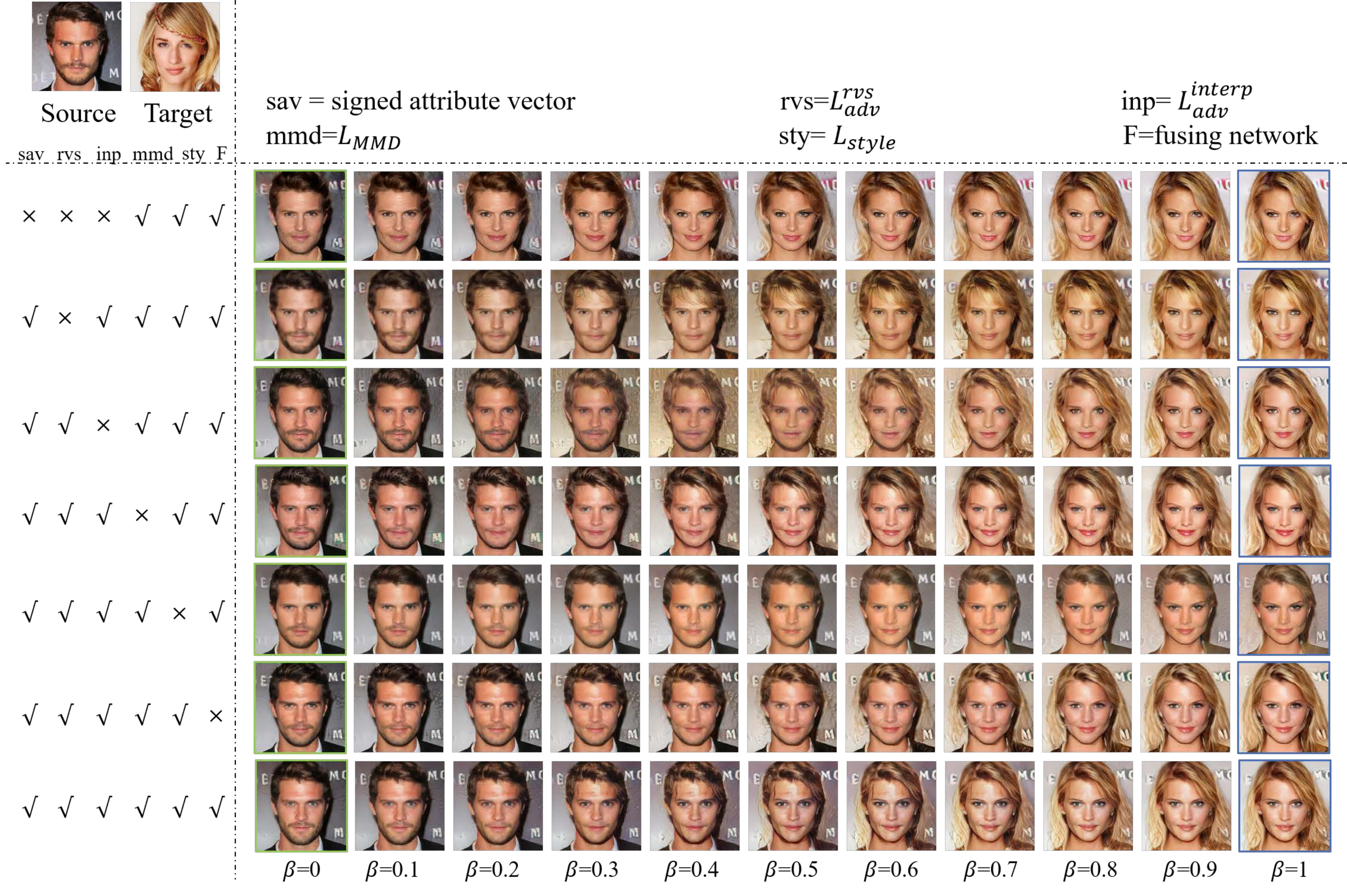}}
\end{minipage}

   \begin{minipage}{1\textwidth}
    \subfloat[Quantitative comparisons]{
    \includegraphics[width=\linewidth]{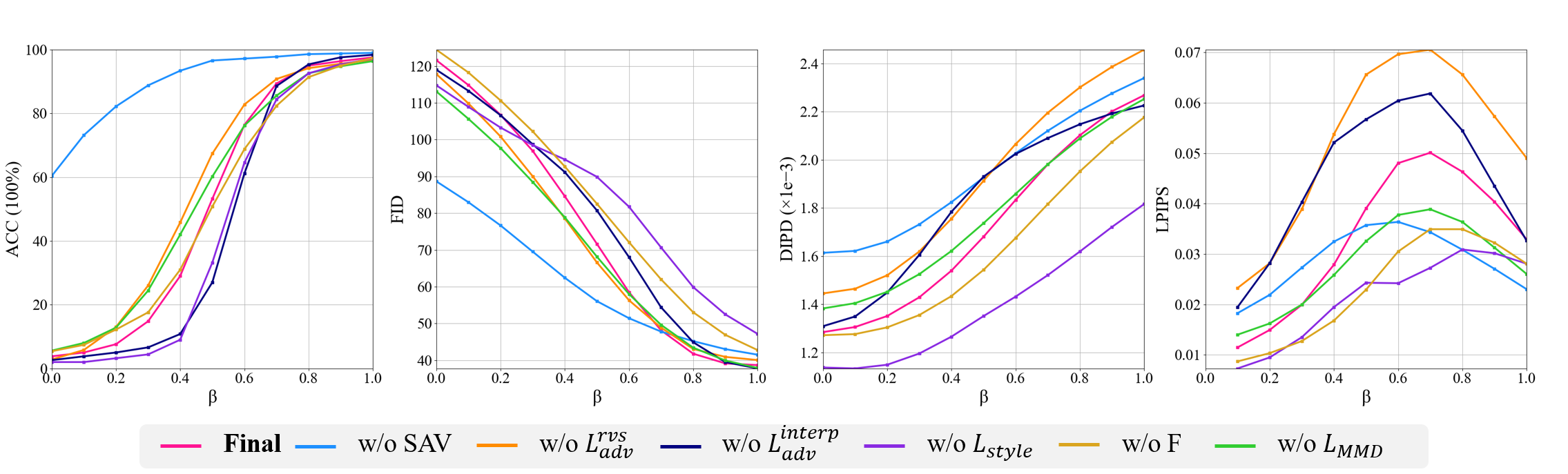}
    }
    \end{minipage}
   
    \caption{\textbf{Ablation study on the male$\to$female translation.}
(a) Green and blue bounding boxes denote generated images of the source and target domain, respectively.
   (b)  The x-axis is the $\beta$. From left to the right, the y-axis in each sub-figure are target domain translation ACC (the larger, the better), target domain FID (the smaller, the better), DIPD (the smaller, the better), and LPIPS between two adjacent interpolated images (the smaller, the better), respectively.
    }
    \label{fig:ablation_study}
\end{figure*}

\subsubsection{User Study}
We evaluate the user preference among the proposed method and the state-of-the-art approaches through pairwise comparisons on the CelebA-HQ and AFHQ datasets.
For each test, we present a source image and two videos with translation images by the proposed and other methods and some examples of the target domain.
For each participant, we randomly select four images from the test set of the CelebA-HQ and six images from the test set of the AFHQ dataset for each evaluated method.
To better present intermediate translated results, we generate videos by repeating each interpolated image five times.
We then ask three questions for each test: (1)
Which model translates images better (both in the transition and the end) in terms of realism? (2) Which method generates smoother translated images? (3) Overall, which approach performs better?

We collect the answers from $20$ participants.
\tabref{user_study} shows that most participants prefer the results generated by the proposed method (from $83.75\%$to $97.50\%$ on the CelebA-HQ dataset and $64.17\%$ to $98.33\%$ on the AFHQ dataset) than those by all the other evaluated approaches.
For image realism, we analyze that users prefer more considerable variation, \eg exhibiting hairstyle variation in the male $\rightleftharpoons$female translation.
Although we ask users to pay attention to the intermediate results, we find that participants are still more concerned with the final translated effects than the intermediate results.
Thus, the gap between StarGAN-v2 (or FUNIT) and the proposed method is smaller than that using the interpolated domain label for interpolation. 
When judging the translated images in terms of smoothness, it is difficult for participants to ignore the influence of other factors such as translation efficiency.
For example, \figref{compare_result1} and \figref{compare_result2} show that the HomoInterpGAN and Fader Networks methods can translate images with a smooth transition but not to the target domain. 
However, most subjects still prefer the proposed method in terms of smooth translation, as shown in \tabref{user_study}.

\subsection{Ablation Studies}
To better understand each component's effectiveness in the proposed method, we present the ablation studies on the male$\to$female translation in \figref{ablation_study}.
We first analyze three proposed components: SAV, reverse sign domain adversarial loss $\mathcal{L}^{\mathrm{rvs}}_\mathrm{{adv}}$, and interpolated domain adversarial loss $\mathcal{L}^\mathrm{interp}_\mathrm{adv}$.
Then, we demonstrate the efficiency of the style reconstruction loss $\mathcal{L}_\mathrm{style}$, the fusing network $F$, and the MMD constraint $\mathcal{L}_\mathrm{mmd}$.

\Paragraph{Proposed components.}
Both quantitative and qualitative results demonstrate that the proposed SAV plays an essential role in the continuous translation across domains.
Without the sign operation, all interpolation results belong to the female domain, as illustrated in \figref{ablation_study}(a).
\figref{ablation_study}(b) shows that,  compared to the final model, most translated images are classified to the female domain when $\beta <0.5$ (ACC $> 50\%$).
Since the dataset contains more female images than male images in the training set (17K vs. 9K), the model tends to learn more female domain attributes without the sign information embedding.
The $\mathcal{L}^{\mathrm{rvs}}_\mathrm{{adv}}$ ensures the domain-membership of the translated images using the sign-symmetrical attribute vector.
As shown in \figref{ablation_study}(a),
without $\mathcal{L}^{\mathrm{rvs}}_\mathrm{{adv}}$, the translated target image cannot preserve the pose of the source image well.
In particular, the pose of interpolated faces images varies from right to left.
Thus, it achieves the highest DIPD scores when $\beta > 0.5$ and most variation in the LPIPS score.
When the model is trained without $\mathcal{L}^\mathrm{interp}_\mathrm{adv}$, 
it achieves the third-highest FID score when $\beta > 0.5$ and second-largest variation in the LPIPS score.
Furthermore, \figref{ablation_study} shows that it does not generate the intermediate results between two domains well.
Therefore, applying $\mathcal{L}^\mathrm{interp}_\mathrm{adv}$ of interpolated results on the trajectory between sign-symmetrical attribute vectors is essential to improve the quality and smoothness.

\Paragraph{Style reconstruction loss and fusing network.}
We observe that training without $\mathcal{L}_\mathrm{style}$ cannot capture the style of the reference image for the translated image, as presented in \figref{ablation_study}(a).
Therefore, it achieves the highest FID values when $\beta >0.5$.
Feeding the attribute vector into a fusing network before the generator enhances the quality of interpolated results, as shown in the ``FID vs. $\beta$'' curve of \figref{ablation_study}(b).

\begin{figure}[!t]
\centering
    \includegraphics[width=\linewidth]{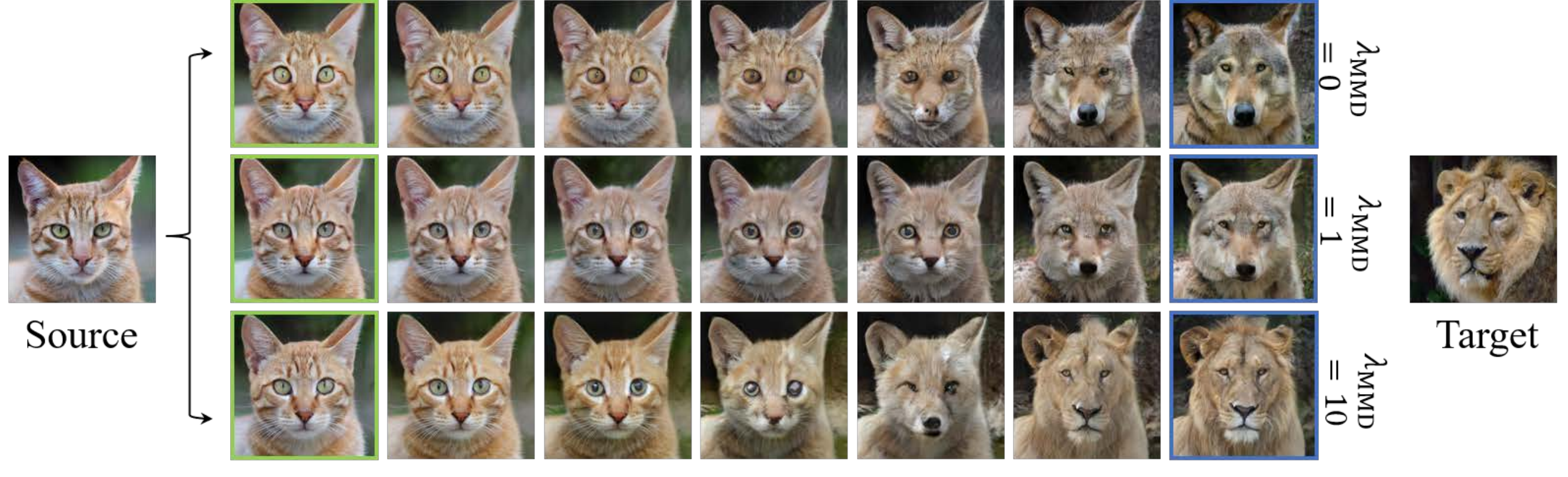}
    \caption{ \textbf{Ablation study on $\mathcal{L}_{\mathrm{MMD}}$ with different values of $\lambda_{\mathrm{MMD}}$ on the Cat $\to$ Wildlife translation.}
Green and blue bounding boxes denote generated images of the source and target domain, respectively.
    }
    \label{fig:MMD}
\end{figure}

\Paragraph{MMD constraint.}
The variation of the ``LPIPS vs. $\beta$'' curve from the model trained without $\mathcal{L}_{\mathrm{MMD}}$ in the male$\to$female translation is smoother than that by the final model.
The translation ACC, FID as well as DIPD scores also have comparable performance against the final model.
However, the translated images do not capture the style of lions in the AFHQ dataset when the weighting parameter of $\lambda_{\mathrm{MMD}}$ is small, as shown in \figref{MMD}.
Since the wildlife domain contains numerous species such as the lion, tiger, fox, and wolf, the attribute vector extracted by $E_a$ can better embed the sign information and further represent the style information of target images by aligning the distribution under the constraint of $\mathcal{L}_\mathrm{MMD}$.
Thus, we use this constraint in the final model.

\begin{figure}[!t]
\centering
\includegraphics[width=\linewidth]{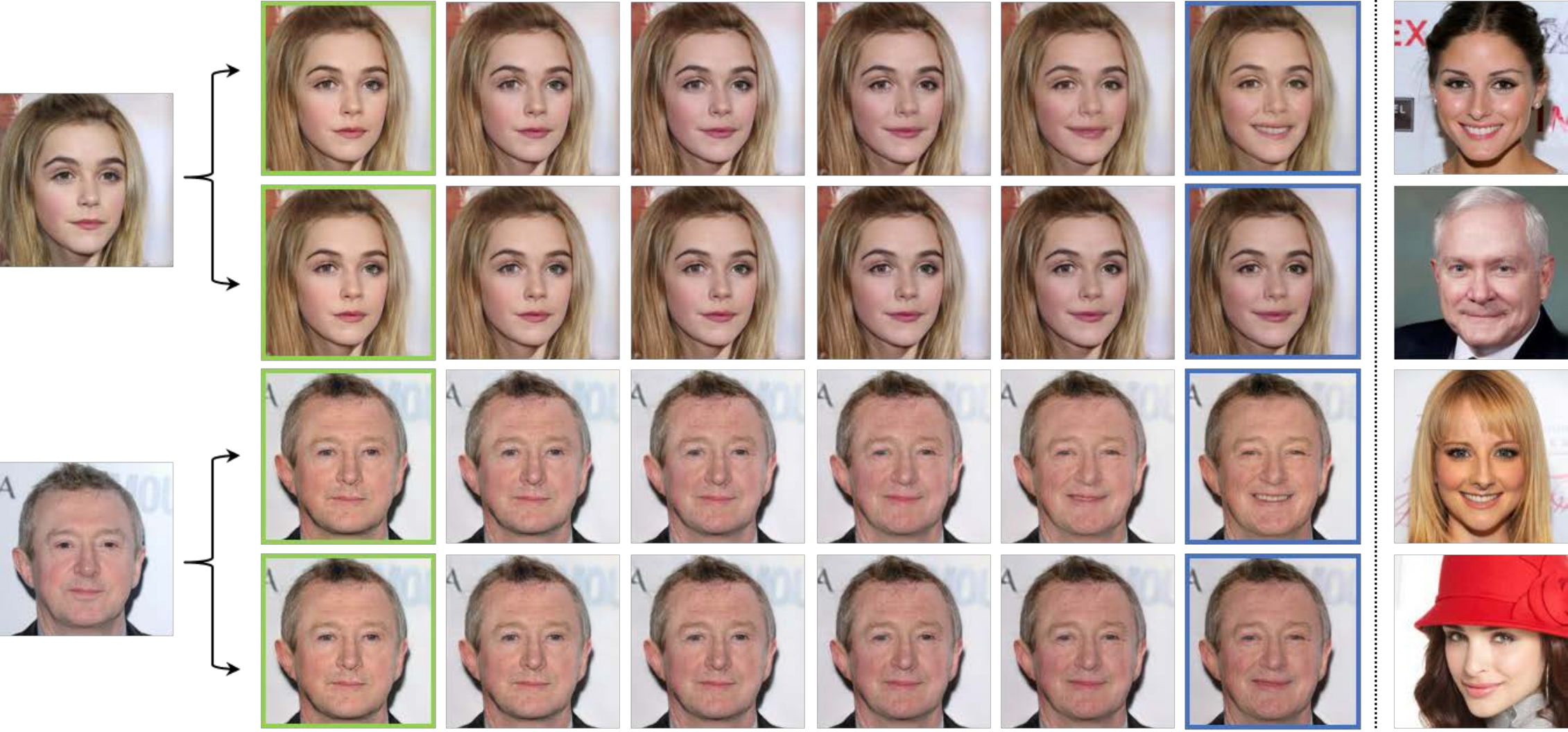}
\mpage{0.1}{Source}\hfill\mpage{0.7}{\qquad Interpolated results 
}\hfill\mpage{0.1}{Target}
\caption{ \textbf{Continuous expression translation.}
Source domain: No-Expression; Target domain: With-Expression.
We present more results in the supplementary material.
  }
    \label{fig:expression}
\end{figure}
\begin{figure}[!t]
\centering
\includegraphics[width=\linewidth]{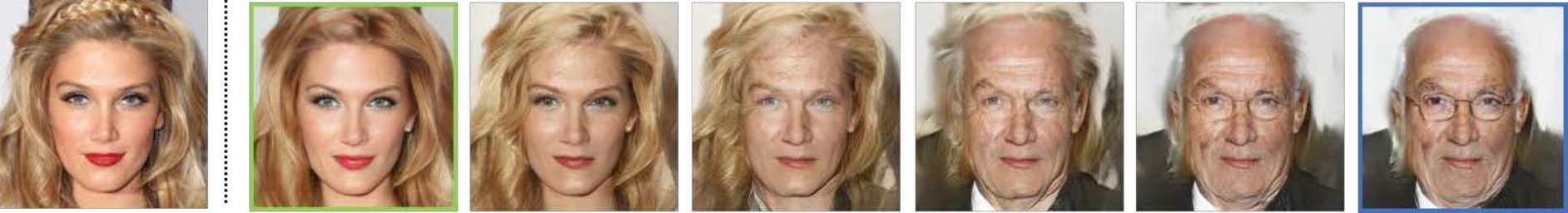}
\mpage{0.1}{Source}\hfill\mpage{0.85}{Interpolated results}
\caption{ \textbf{A limitation case.}
Our method manipulates domain-independent attributes such as age and eyeglasses in the female$\to$male translation.
  }
    \label{fig:limitation}
\end{figure}

\subsection{Discussion}
\Paragraph{Applicability analysis.}
We conduct continuous translation on facial expression to demonstrate the proposed method’s applicability on ``shape-like'' attributes manipulation. 
Similar to~\cite{chen2019homomorphic}, we split the CelebA-HQ dataset into two domains using attributes related to the expression, \ie``Smile'' and ``Mouth-slightly open''.
\figref{expression} shows that our model can generate diverse and continuous expression translation results.
These results show that the proposed model 
is task-independent and can learn domain-specific attributes based on the datasets and tasks. 

\Paragraph{Limitation.} 
There are several limitations of the proposed model. 
Without fine-grained annotations, some domain-independent attributes such as age and eye-glasses would leak into the attribute latent space in the male$\rightleftharpoons$female translation task, as illustrated in \figref{limitation}.
Although the proposed framework can continuously translate a source image into the target domain using the proposed signed attribute vectors, it is of great interest to derive such fine-grained visual information directly from images via disentangled representations.
Our future work will focus on developing effective representation learning schemes to facilitate smooth image translation.

\section{Conclusions}
\label{sec:conclusion}
In this paper, we present a signed attribute vector to enable continuous and diverse I2I translation across domains.
To enhance the continuous translation quality, we propose to use the sign-symmetrical attribute vectors to form a translation trajectory between different domains.
Then, we leverage the domain information of intermediate results for adversarial training.
We evaluate our methods on a wide range of I2I translation tasks.
Both qualitative and quantitative results demonstrate that the proposed method achieves high-quality and diverse continuous translation across domains.

\bibliographystyle{spmpsci}      
\bibliography{ijcv19}  
\end{document}